\documentclass[journal,twoside,web]{ieeecolor}
\usepackage{colortbl}

\usepackage{tmi}
\usepackage{cite}
\usepackage{amsmath,amssymb,amsfonts}
\usepackage{arydshln}
\usepackage{multirow}
\usepackage[noend]{algpseudocode}
\usepackage{algorithmicx,algorithm}
\usepackage{graphicx}
\usepackage{textcomp}
\usepackage{epstopdf}
\usepackage{booktabs}
\usepackage[section]{placeins}
\usepackage{stfloats}
\usepackage{bm}
\usepackage{xstring}
\usepackage{bbding}
\usepackage{color}
\usepackage{ulem}
\definecolor{darkcyan}{RGB}{0,138,218}
\usepackage[colorlinks=true, allcolors=darkcyan,citecolor=blue]{hyperref}
\usepackage[switch]{lineno}
\useunder{\uline}{\ul}{}
\usepackage{colortbl}
\definecolor{lightgray}{gray}{0.84}

\def\BibTeX{{\rm B\kern-.05em{\sc i\kern-.025em b}\kern-.08em
	T\kern-.1667em\lower.7ex\hbox{E}\kern-.125emX}}

\begin{document}
\title{AttriPrompter: Auto-Prompting with Attribute Semantics for Zero-shot Nuclei Detection via Visual-Language Pre-trained Models.}
\author{Yongjian Wu, Yang Zhou, Jiya Saiyin, Bingzheng Wei, Maode Lai, Jianzhong Shou, and Yan Xu
	\thanks{This work is supported by the National Natural Science Foundation in China under Grant 62371016, U23B2063, and 62022010, the Bejing Natural Science Foundation Haidian District Joint Fund in China under Grant L222032, the Fundamental Research Funds for the Central University of China from the CCSE in Beihang University in China, the 111 Proiect in China under Grant B13003. \textit{(Yongjian Wu and Yang Zhou contributed equally to this work. Corresponding author: Yan Xu.)}}
	\thanks{Y. Wu, Y. Zhou, Jiya Saiyin, and Y. Xu are with the School of Biological Science and Medical Engineering, Key Laboratory of Biomechanics and Mechanobiology of Ministry of Education, Beijing Advanced Innovation Center for Biomedical Engineering, Beihang University, Beijing 100191, China (e-mail: xuyan04@gmail.com).}
	\thanks{B. Wei is with ByteDance Inc., Beijing 100098, China.} 
	\thanks{Maode Lai is with Department of Pathology, School of Medicine, Zhejiang University, Zhejiang Provincial Key Laboratory of Disease Proteomics and Alibaba-Zhejiang University Joint Research Center of Future Digital Healthcare, Hangzhou 310053, China.} 
    \thanks{Jianzhong Shou is with the Department of Urology, National Cancer Center, National Clinical Research Center for Cancer, Cancer Hospital, Chinese Academy of Medical Sciences and Peking Union Medical College, Chaoyang District, Beijing 100021, China.} 
 }
\maketitle
\begin{abstract}
Large-scale visual-language pre-trained models (VLPMs) have demonstrated exceptional performance in downstream object detection through text prompts for natural scenes. However, their application to zero-shot nuclei detection on histopathology images remains relatively unexplored, mainly due to the significant gap between the characteristics of medical images and the web-originated text-image pairs used for pre-training. This paper aims to investigate the potential of the object-level VLPM, Grounded Language-Image Pre-training (GLIP), for zero-shot nuclei detection. Specifically, we propose an innovative auto-prompting pipeline, named AttriPrompter, comprising attribute generation, attribute augmentation, and relevance sorting, to avoid subjective manual prompt design. AttriPrompter utilizes VLPMs' text-to-image alignment to create semantically rich text prompts, which are then fed into GLIP for initial zero-shot nuclei detection. Additionally, we propose a self-trained knowledge distillation framework, where GLIP serves as the teacher with its initial predictions used as pseudo labels, to address the challenges posed by high nuclei density, including missed detections, false positives, and overlapping instances. Our method exhibits remarkable performance in label-free nuclei detection, outperforming all existing unsupervised methods and demonstrating excellent generality. Notably, this work highlights the astonishing potential of VLPMs pre-trained on natural image-text pairs for downstream tasks in the medical field as well. Code will be released at \href{https://github.com/wuyongjianCODE/AttriPrompter}{github.com/AttriPrompter}.
\end{abstract}


\section{Introduction}
\label{sec:introduction}


\IEEEPARstart{I}{n} the field of medical image processing, nuclei detection on Hematoxylin and Eosin (H\&E)-stained images holds great importance. It plays a crucial role in the diagnosis of numerous diseases and serves as the foundation for various subsequent analyses, including prognosis evaluation and treatment planning \cite{gleason1992histologic}. To address the aforementioned task, researchers have put forth numerous fully supervised methods \cite{mahanta2021ihc,graham2019hover,yi2019multi}. However, compared to most natural objects, nuclei in H\&E images exhibit high density, meaning they are spatially close to each other \cite{zhou2023cyclic}. Nuclei are often clustered, making it hard to locate all target nuclei without omission. Therefore, the annotation process for detection requires specialized knowledge and is labor-intensive, expensive, and error-prone. To address the aforementioned challenges, researchers have shifted their focus to unsupervised learning \cite{mouelhi2018fast,sahasrabudhe2020self, le2022unsupervised}. However, current unsupervised methods tend to exhibit strong empirical characteristics, meaning users have to set certain parameters based on subjective experience, or the methods are fundamentally based on subjective experience. For example, thresholding-based methods \cite{jiao2006improved,mouelhi2018fast} require setting threshold values based on experience. Self-supervised methods like SSNS \cite{sahasrabudhe2020self} subjectively assume that classifying magnification levels can help identify nuclei size and texture. Some domain adaptation-based methods \cite{le2022unsupervised} include empirical design conditions when aligning source and target domains, as seen in PDAM \cite{liu2020pdam} and DARCNN \cite{hsu2021darcnn}. These methods inevitably introduce subjective biases in the model design process. 


Currently, large-scale visual-language pre-trained models (VLPMs) achieve semantically-rich, versatile, and transferable visual features by learning aligned text and image features from extensive internet-derived text-image pairs \cite{radford2021learning}, and thus have given rise to new unsupervised learning methods. VLPMs have been effectively applied in various tasks like image manipulation, captioning, view synthesis, and object detection in a zero-shot manner \cite{patashnik2021styleclip, li2022blip, jain2021putting, li2022grounded}. Notably, the Grounded Language-Image Pre-training (GLIP) model shows impressive zero-shot capabilities in object detection and phrase grounding \cite{li2022grounded}. Recent studies have shifted focus to VLPMs' applicability in medical imaging \cite{liu2021medical,lu2023visual,wu2023zeroshot}. In light of the above discussions, this paper aims to establish an unsupervised zero-shot nuclei detection system based on VLPMs.

The crucial aspect of applying VLPMs to histopathology image nuclei detection in a zero-shot manner is prompt design. Specifically, due to the domain gap between web-originated text-image pairs of the pre-training data and medical images, VLPMs lack prior knowledge about medical concept words \cite{qin2022medical}. This lack leads to challenges in prompt design for nuclei detection via object-level VLPMs like GLIP. Some researches indicate that constructing appropriate attribute texts enables VLPMs to recognize previously unseen concept words \cite{li2022grounded,yamada2022lemons}. However, predominant approaches involve manually designing attributes, which is laborious, subjective, and prone to significant biases in attribute selection. Qin \textit{et al.} addressed this by designing MIU-VL to automatically generate prompts \cite{qin2022medical}. Nonetheless, their method did not account for H\&E nuclei detection. Retrieving semantically comprehensive attributes from unlabeled image data and automatically constructing optimal prompt sequences for zero-shot nuclei detection utilizing VLPMs remains underexplored. Therefore, in this paper, we conduct an empirical analysis on prompt design for histopathology nuclei detection for the first time based on GLIP. Our key observations reveal that semantic descriptions with finer granularity, especially those incorporating attribute degree descriptions, i.e. the qualitative or quantitative specifications that describe the extent or level of certain attributes of an object, significantly enhance VLPMs' ability to detect unseen concepts. Additionally, we find that arranging prompts with higher relevance at the forefront positively influences GLIP's prediction accuracy.


Building upon these insights, we construct an auto-prompting pipeline using VLPMs, named AttriPrompter, which mainly comprises three steps: attribute generation, attribute augmentation, and relevance sorting. Initially, attribute generation utilizes GLIP and the vision-language understanding model BLIP \cite{li2022blip} to create attribute words describing nuclei shape and color. Next, attribute augmentation which includes synonym augmentation and degree augmentation is conducted by employing the pre-trained GPT model \cite{radford2019language}. Notably, degree augmentation contributes to acquiring attribute words with finer descriptive granularity such as the depth or lightness of a color and the degree of curvature in shapes, which are crucial for nuclei recognition. Finally, in relevance sorting, attribute words and medical nouns are combined to form detection prompts and then listed in descending order according to the relevance, creating the final prompt sequence. AttriPrompter utilizes VLPMs' inherent text-to-image alignment, generating text prompts that are rich in attribute semantics. This sequence empowers VLPM to localize unfamiliar nuclei, even with unseen medical images, and serves as a paradigm for VLPM's zero-shot transfer in other medical image modalities. This sequence is input into the object-focusing VLPM GLIP for initial zero-shot nuclei detection.


However, due to the higher object density of nuclei, the established architecture of GLIP still faces challenges when detecting dense and complex nuclei, leading to missed detections, false positives, and overlapping. Thus, to relax from the architectural restriction while maintaining the generalizable knowledge learned by GLIP, a self-trained knowledge distillation framework is proposed. In this framework, GLIP serves as the teacher network with the initial predictions as pseudo labels, while an advanced detection-specific network is adopted as the student network which is trained on these pseudo labels. We optimize the student network's feature-extracting backbone using an auxiliary knowledge distillation loss \cite{wang2021knowledge} to prevent forgetting GLIP's powerful object-level knowledge in a self-training manner \cite{dopido2013semisupervised}. The self-trained knowledge distillation framework effectively leverages GLIP's potent prior knowledge, mitigating knowledge forgetting \cite{chen2019catastrophic} and enhancing zero-shot detection efficacy.

Comprehensive experiments demonstrate that our zero-shot nuclei detection framework achieves superior performance compared to other unsupervised methods, including concurrent open-vocabulary VLPM-based approaches. The generality of the framework has also been validated across various datasets and detection architectures. Our contributions include: 
\begin{itemize}
    \item Proposing a novel zero-shot nuclei detection method based on VLPMs for the first time, outperforming comparison unsupervised methods and demonstrating excellent generality.
    \item Developing a new auto-prompting pipeline, named AttriPrompter, comprising attribute generation, attribute augmentation, and relevance sorting, to generate granular prompts that are semantically aligned with nuclei.
    \item Devising a self-trained knowledge distillation framework to address high nuclei density. It significantly reduces missed detections, false positives, and overlapping.
\end{itemize}

This paper serves as a substantial extension of our previously published work \cite{wu2023zeroshot}. The key enhancements include: (1) Detailed analysis on prompt design for nuclei detection on H\&E images. (2) Proposing a new prompt generation method, which utilizes degree augmentation to generate descriptions with finer granularity based on attribute semantics, and fully exploits the detection potential of prompt sequences through relevance sorting. (3) Employing knowledge distillation to effectively leverage the powerful prior knowledge of GLIP and mitigate knowledge forgetting. (4) More extensive experiments of the proposed framework are conducted, including additional comparison methods, new validation datasets, cross-domain experiments, and semi-supervised copmarison, etc.


\section{Related works}

\subsection{Nuclei Detection} Deep learning methods have recently dominated the area of nuclei detection in histopathology analysis, outperforming conventional approaches \cite{gleason1992histologic, irshad2013methods, mouelhi2018fast}. Various fully-supervised networks and training frameworks tailored for nuclei detection or instance segmentation on histopathology images have been proposed \cite{mahanta2021ihc,graham2019hover,yi2019multi}. However, researchers are increasingly shifting focus from fully-supervised to weakly-supervised \cite{li2019signet, qu2020weakly} and unsupervised approaches \cite{sahasrabudhe2020self, le2022unsupervised} to overcome annotation costs and efficiency challenges. Sahasrabudhe \textit{et al.} present a self-supervised approach for segmentation of nuclei for whole slide histopathology images based on attention mechanism \cite{sahasrabudhe2020self}. Hsu \textit{et al.} propose DARCNN, which adapts knowledge of object definition from COCO \cite{lin2014microsoft} to multiple biomedical datasets including histopathology images \cite{hsu2021darcnn}. However, these unsupervised methods often rely on empirical design choices and can introduce subjective biases during the model design process.

\subsection{Visual-Language Pre-trained Models}\; Visual-language pre-trained models (VLPMs), trained on extensive text-image pairs from the web, use separate encoders to align text and image features \cite{chen2023vlp}. These models serve various purposes, including representative learning \cite{radford2021learning} and image-to-text generation \cite{li2022blip}. VLPMs are often based on contrastive learning \cite{jaiswal2020survey,radford2021learning}. Among them, Contrastive Language-Image Pre-training (CLIP) \cite{radford2021learning} stands out as a prominent work, producing models capable of generating semantic-rich, general, and transferable image-level features. These features allow for direct applications in downstream visual recognition tasks through transfer learning, yielding impressive results \cite{patashnik2021styleclip,li2022blip,zhou2022extract,zhao2022exploiting,lin2022learning}. The medical imaging field has also benefited from CLIP, with notable improvements in relevant tasks \cite{eslami2021does,liu2023clip}. Researchers have endeavored to extend CLIP's image-level knowledge to dense prediction tasks such as semantic segmentation \cite{zhou2022extract} and object detection \cite{zhao2022exploiting,lin2022learning}. However, when faced with complex and densely packed content in images, these methods have exhibited poor performance. Recently, Grounded Language-Image Pre-training (GLIP), a novel approach that pre-trains on grounding text-image data, has outperformed CLIP in object-level tasks \cite{li2022grounded}. It even rivals fully-supervised counterparts in zero-shot object detection and phrase grounding tasks. In light of these developments, the present study aims to explore whether the knowledge obtained from VLPMs can be leveraged to handle more challenging nuclei detection tasks effectively.

\subsection{Prompt Design}\; Prompt design, originating from natural language processing, involves techniques to provide suitable prompts for adapting pre-trained language models (LM) to downstream tasks \cite{liu2023pre}. These techniques are also known as prompt engineering. Concretely, the model's input is augmented with an additional prompt, which can be in the form of manually created natural language instructions \cite{wei2022chain}, automated generated natural language instructions \cite{zhang2022automatic}, or automated generated vector representations \cite{lester2021power}. With suitable prompts, the pre-trained LM can predict the desired output, sometimes even without any additional task-specific training \cite{brown2020language}. Similarly, for VLPMs, manually designed prompts can be used as input for the text encoder, enabling downstream visual tasks through pre-trained text-to-image alignment \cite{gu2023systematic}. For instance, CLIP utilizes questions as prompts to perform classification tasks \cite{radford2021learning}, while GLIP concatenates the detected object nouns as prompts for object detection \cite{li2022grounded}. However, manual prompt design can be tedious and subjective, and may introduce significant bias. To address this, some studies have attempted to set learnable prompts in few-shot settings to avoid manual design, such as CoOp \cite{zhou2022learning} and VPT \cite{jia2022visual}. Nevertheless, these approaches still require labeled data for training. Recently, Yamada \textit{et al.} revealed that constructing prompts using attribute adjectives can aid VLPMs in recognizing unfamiliar object concepts \cite{yamada2022lemons}. Building on this, Qin \textit{et al.} developed MIU-VL, an automated prompt generator, although it did not cater to H\&E nuclei detection \cite{qin2022medical}. This paper proposes a new auto-prompting method called AttriPrompter for zero-shot nuclei detection utilizing existing VLPMs, which is based on attribute semantics and achieves better results.

\subsection{Knowledge Distillation}\; Knowledge distillation (KD) has been extensively employed as a constructive method to transfer information from one network to another during training \cite{hinton2015distilling,wang2021knowledge}. KD has found wide-ranging applications in model compression \cite{buciluǎ2006model} and knowledge transfer \cite{ahn2019variational}. Recently, the research community has been captivated by the remarkable potential demonstrated by VLPMs. As a natural progression, researchers have turned their attention to KD, to transfer VLPMs' knowledge more effectively to downstream tasks \cite{zhang2023vision}, such as semantic segmentation \cite{ding2022decoupling,luddecke2022image} and object detection \cite{zhao2022exploiting,lin2022learning,wang2023object}, to achieve superior results. Inspired by these developments, we introduce self-trained knowledge distillation, based on self-training \cite{dopido2013semisupervised}. This approach utilizes GLIP's powerful prior knowledge, effectively mitigating the issue of knowledge forgetting in nuclei detection tasks.


\section{Preliminary: Grounded Language-Image Pre-training Models}




VLPMs adopt extensive web-originated text-image pairs $\{ (x,t )\}_{i}$ for contrastive learning to attain the text-to-image alignment over the text and visual embeddings. The loss function for the learning process can be formularized as: 
\begin{equation}
\label{eq1}
\mathcal{L}_c=-\sum_{i=1}^K \log \frac{\exp \left(\cos \left( E_{\mathcal{V}}\left(x_i\right) \cdot E_{\mathcal{T}}\left(t_i\right)\right) \right)}{\sum_{j=1}^K \exp \left(\cos \left(E_{\mathcal{V}}\left(x_i\right) \cdot E_{\mathcal{T}}\left(t_j\right)\right)\right)},
\end{equation}
among which, $E_{\mathcal{V}}$ and $E_{\mathcal{T}}$ denote the image and text encoder respectively, $K$ represents the batchs size in one step, $\cos(\cdot)$ denotes cosine similarity. Li \textit{et al.} introduced the Grounded Language-Image Pre-training (GLIP) models \cite{li2022grounded}, which enhance object representation over basic VLPMs. GLIP's image encoder produces visual embeddings for image regions corresponding to each object word in the text, using web-originated phrase grounding pairs to enrich text encoder object concepts. It also employs a deep cross-modality fusion with cross attention (X-MHA) \cite{chen2021crossvit} for multi-level alignment. The fusion process achieves a more nuanced alignment of text embedding $P$ and visual embedding $R$. This process enables GLIP to achieve enhanced semantic aggregation at the object level, outperforming traditional VLPMs in object-level zero-shot transfer. Consequently, we utilize a pre-trained GLIP model for developing the label-free nuclei detection system.

\section{Prompt analysis for nuclei detection}
\label{sec:promptana}
Prompt design is crucial for applying VLPM to zero-shot histopathology image nuclei detection. With the text-to-image alignment pre-training, GLIP is able to perform zero-shot detection directly by using a text prompt. In natural scences, the default prompt for the detection task proposed by Li \textit{et al.} is the concatenation of object nouns, such as ``object noun 1. object noun 2. …,'' \cite{li2022grounded}. However, since GLIP's semantic space exhibits limited coverage of medical terms, the mere concatenation of nouns fails to facilitate effective nuclei retrieval. Inspired by previous findings that semantic descriptions can enhance object detection \cite{li2022grounded,yamada2022lemons}, we believe that by constructing appropriate attribute descriptions, VLPMs can achieve label-free detection of unseen medical objects. Initially, a prompt analysis based on manual design is conducted on the MoNuSeg \cite{kumar2017dataset} test dataset with GLIP-L to investigate optimal prompt designs that are conducive to nuclei detection. Implementation details are shown in \autoref{experiments}.

We observed that the accuracy of GLIP's predictions increases as the prompt descriptions exibit finer granularity, i.e. become more detailed and precise, as shown in \autoref{tab:PAdegree}. Beginning with ``Nuclei'', the bounding boxes are progressively honed to greater accuracy as the semantic details are incrementally enhanced. Particularly, GLIP can produce more accurate results when degree adverbs are incorporated into the prompt. It demonstrates the critical role of degree descriptions in recognizing unseen concepts with GLIP, an aspect overlooked in previous works. 

Additionally, it was noted that the prompt sequences' order impacts GLIP's performance. \autoref{fig:parelevance} depicts this, showing the average relevance as the mean cosine similarity between GLIP-generated text embeddings and visual embeddings of images in the MoNuSeg training set (details in \ref{sec:RS}). An example image from the testing set is presented in \autoref{fig:parelevance}, with mAP calculated across the entire Testing Set. Results indicate that positioning higher relevance prompts earlier in the sequence positively influences GLIP, which further enhances the accuracy of zero-shot prediction.

\begin{figure*}[t]
\includegraphics[width=\textwidth]{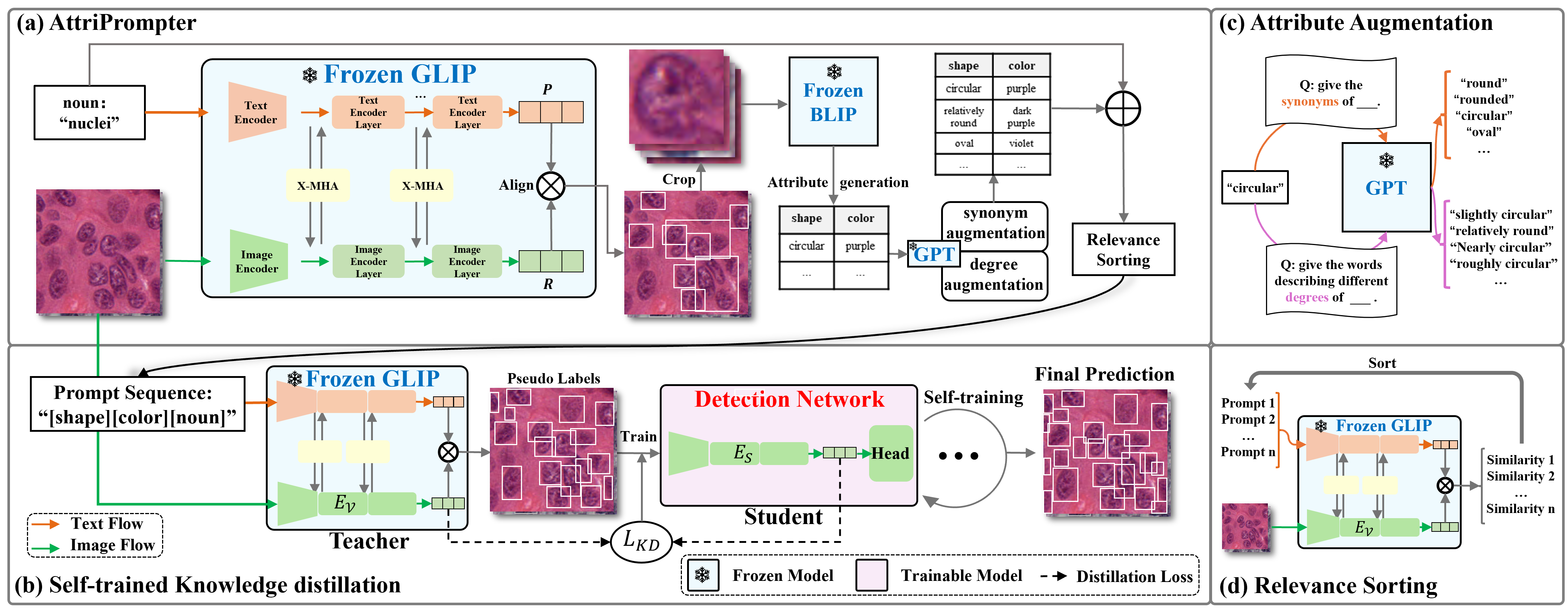}
\caption{Method illustration. \textcolor{darkcyan}{(a)} AttriPrompter: utilizing GLIP \cite{li2022grounded} and BLIP \cite{li2022blip} for attribute generation; exploiting GPT model \cite{radford2019language} for attribute augmentation; ending with relevance sorting of prompts composed of attribute words and medical nouns. \textcolor{darkcyan}{(b)} Iterative optimization through self-trained knowledge distillation. Panels \textcolor{darkcyan}{(c)} and \textcolor{darkcyan}{(d)} detail the attribute augmentation and relevance sorting processes.} \label{fig1}
\end{figure*}



\section{Approach}
Based on the observations from the previous section, we establish an auto-prompting method, AttriPrompter, to avoid manual design. Additionally, we propose a self-trained knowledge distillation framework to further improve the performance. The overall framework is illustrated in \autoref{fig1}.

\subsection{AttriPrompter}

AttriPrompter consists of three steps: attribute generation, attribute augmentation, and relevance sorting, utilizing established VLPMs to automatically generate suitable prompts for GLIP to detect nuclei in a zero-shot manner.
\subsubsection{Attribute Generation}

\begin{table}[t]
\caption{Mean Average Precision (mAP) on the MoNuSeg Testing Set Using Varied Text Prompts. The prompt descriptions incrementally increase in detail from top to bottom.}
\label{tab:PAdegree}
\begin{center}
\begin{small}
\resizebox{0.75\columnwidth}{!}{
\begin{tabular}{cc}
\toprule[1.5pt]
\textbf{Prompt}                      & \textbf{mAP} \\ \midrule[1.2pt]
``Nuclei.''                              & 0.034        \\
``Round purple nuclei.''                 & 0.046        \\
``Relatively round dark purple nuclei.'' & 0.115        \\ \bottomrule[1.5pt]
\end{tabular}
}
\end{small}
\end{center}
\vskip -0.1in
\end{table}

\begin{figure}[t]
\includegraphics[width=\linewidth]{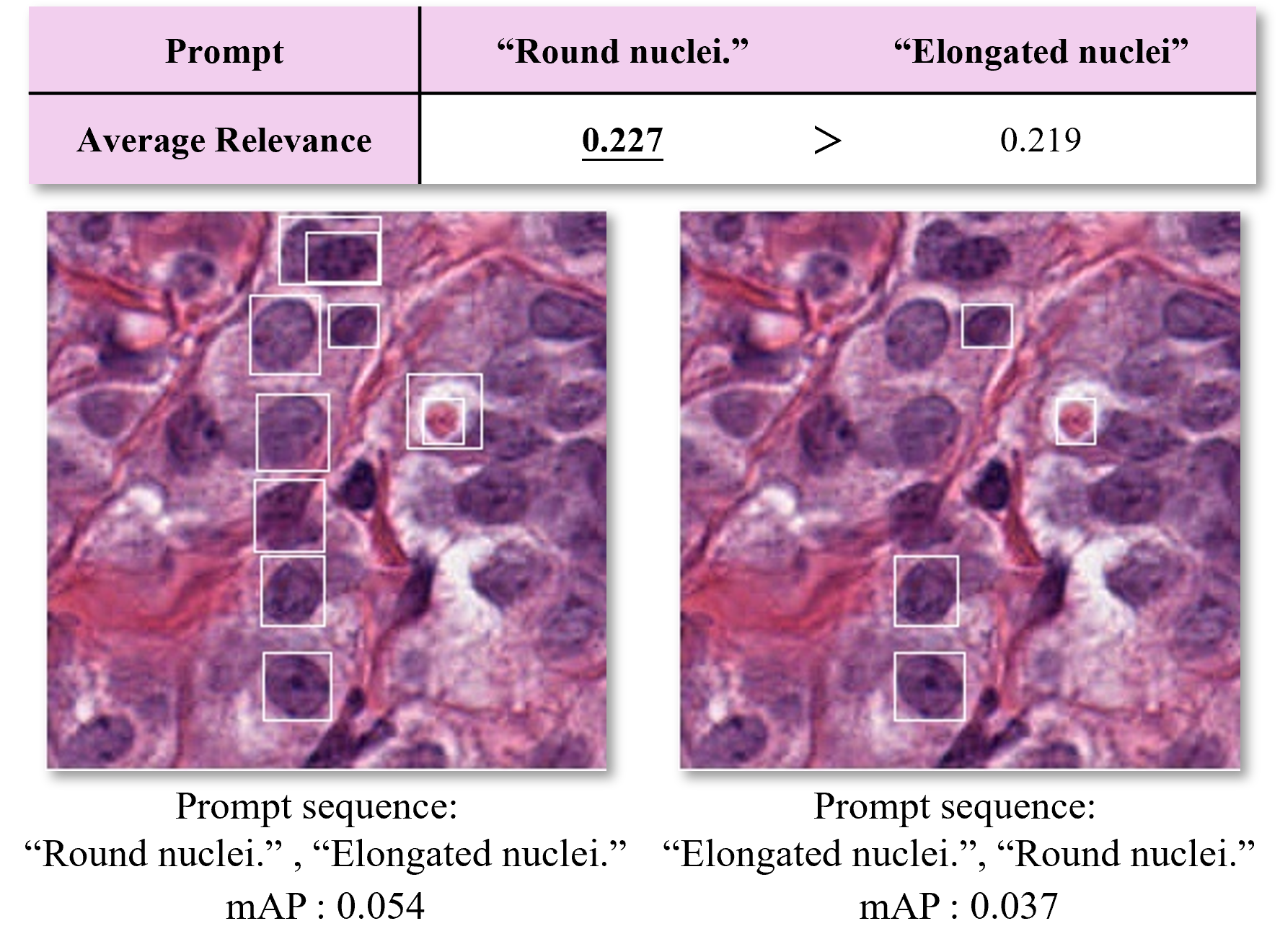}
\caption{Positioning higher relevance prompts earlier in the sequence positively influences GLIP, enhancing prediction accuracy. The boxes are shown in white. The values of mAP are calculated across the entire MoNuSeg testing set.} \label{fig:parelevance}
\end{figure}

\begin{figure}[t]
\includegraphics[width=\linewidth]{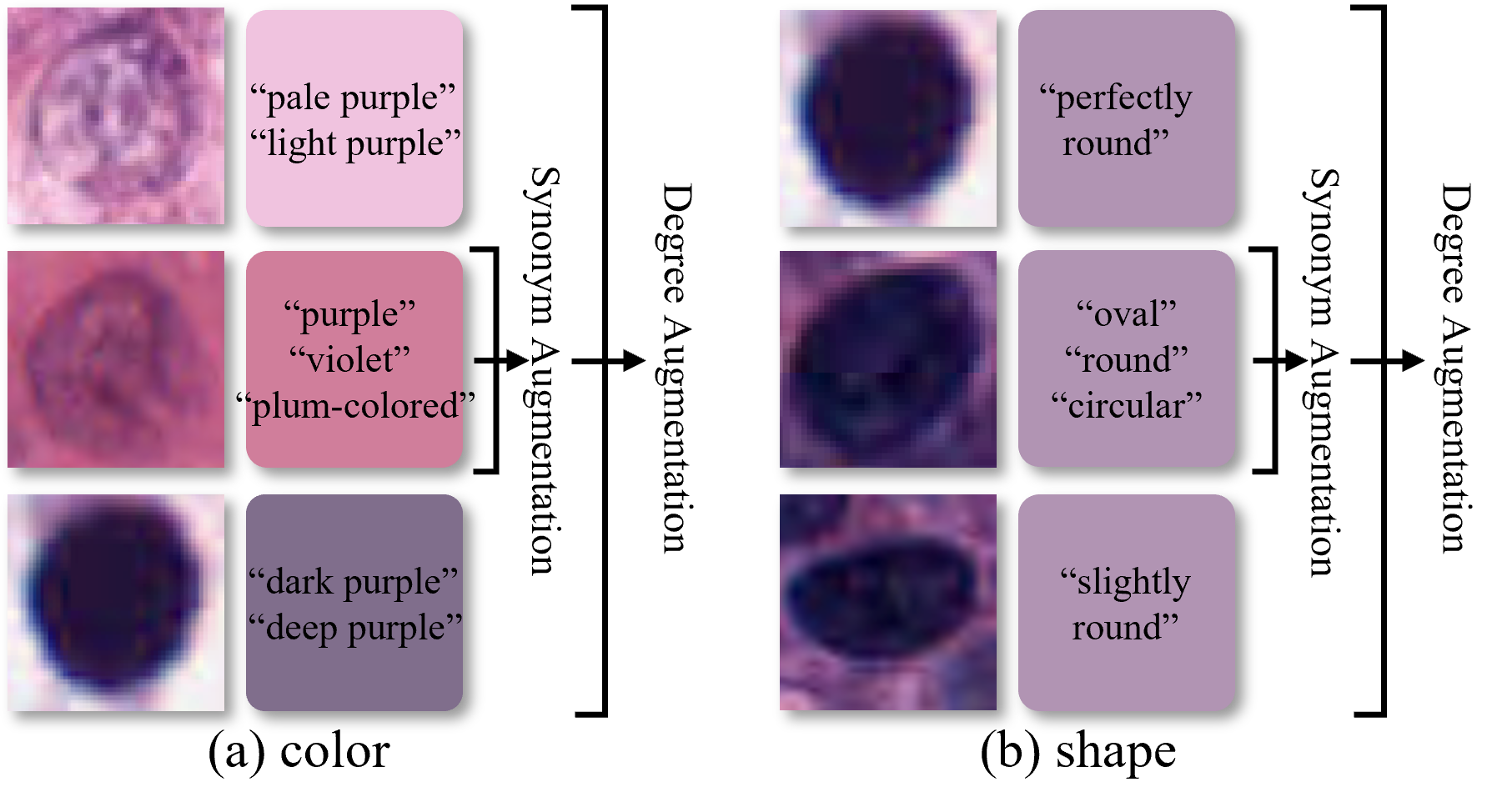}
\caption{Semantic coverage of nuclei with diverse appearances after attribute augmentation. \textcolor{darkcyan}{(a)} Augmentation for color ``purple''. \textcolor{darkcyan}{(b)} Augmentation for shape ``round''. Adjectives within the same frame are synonyms, while those in different frames represent varying degrees of the attribute.} \label{fig:augmentation}
\vskip -0.15in
\end{figure}

BLIP is a vision-language understanding VLPM dedicated to image captioning and describing \cite{li2022blip}. We first utilize BLIP to automatically generate attribute words for describing nuclei, which avoids manual design and ensures the attributes adopted in prompts coordinate with the text-to-image alignment of VLPMs. The process is as follows: Firstly, directly use the target medical nouns as the prompt to obtain coarse predictions with GLIP. Subsequently, crop all the coarse boxes as image patches, which are then input into a fixed BLIP, automatically generating attribute words through visual question answering (VQA) \cite{antol2015vqa}. Word frequency statistics and classification techniques are then employed to obtain words that represent the object's shape and color since they are the most relevant attributes depicting nuclei.

Through these two steps, we leverage off-the-shelf VLPMs to automatically generate appropriate attribute words for nuclei description. This process harnesses the robust text-to-image aligning capabilities of GLIP and BLIP, showcases excellent interpretability, and avoids the time-consuming labor-intensive process of manual design.
\subsubsection{Attribute Augmentation} As shown in \autoref{fig1}.(c), attribute augmentation is composed of synonym augmentation and degree augmentation.

\textbf{Synonym augmentation.} Synonyms can convey the same semantics \cite{he2016automatic}. For a thorough description, the attribute words are augmented with synonyms retrieved by GPT through question answering. GPT is a language model trained on massive text data, which can provide coherent responses to the input text \cite{radford2019language}. Therefore, we adopt a pre-trained GPT model to conduct synonym augmentation, aiming to enhance the descriptive diversity which aids in mitigating the potential instability of prompts and improving the performance of GLIP.

\textbf{Degree augmentation.} Prior research \cite{li2022grounded,yamada2022lemons} and our empirical experiments illustrate that semantic descriptions with finer granularity can facilitate more precise object localization. When ``Nuclei.'' is the prompt, the prediction generated by GLIP is coarse and full of background false positives. Therefore, attribute generation and synonym augmentation may not always yield precise descriptors with proper granularity, such as the depth of color and the degree of curvature in shapes. To enhance the granularity of the semantic descriptions, we introduce degree augmentation to broaden semantic representation. As illustrated in \autoref{fig:augmentation}, degree augmentation generates attributes of different degrees to implement the synonym augmentation, expanding the semantic range and increasing the probability of accurate descriptors appearing in the candidate attribute set. Degree augmentation also adopts the pre-trained GPT model, attaining depictive words by VQA. 

\subsubsection{Relevance Sorting}
\label{sec:RS}
The obtained attribute words are combined with medical nouns to automatically form the triplet detection prompt by the template “[shape][color][noun].”. However, degree augmentation may introduce inaccurate words of low relevance. Since relevance is represented by cosine similarity between visual embedding $R$ and text embedding $P$, we directly utilize cosine similarity for prompt filtering. Prompts and each image $x_{i}$ in the training set are encoded with GLIP to generate $R$ and $P$, then we calculate the mean cosine similarity of a single prompt ${\mathcal{T}}$ with the whole training set, formularized as follows:
\begin{equation}
\label{eq4}
\cos \left(E_{\mathcal{V}}\left(x_i\right), E_{\mathcal{T}}(t)\right)=\frac{1}{n} \sum_{i=1}^n \frac{R \cdot P}{\left\|R\right\|_2 \cdot\left\|P\right\|_2},
\end{equation}
where $n$ denotes the number of training images, $R = E_{\mathcal{V}}\left(x_i\right)$, $P = E_{\mathcal{T}}(t)$. Please note that $R$ represents all candidate boxes, that the index of boxes is omitted for simplicity, and the final similarity is obtained by calculating the average over all boxes. Subsequently, the top $N$ prompts with the highest similarity are selected. According to the previous observation that arranging prompts with higher relevance at the front exhibits a positive feedback effect on GLIP's prediction, we organize these $N$ prompts in descending order based on similarity. Finally, the sorted prompt sequence is fed into GLIP to achieve initial zero-shot nuclei detection. This sorting process enhances the stability of GLIP's output, optimizing the overall detection performance.

\subsection{Self-trained Knowledge Distillation}
Unlike natural images, nuclei in histopathology exhibit high object density \cite{zhou2023cyclic}. Although the grounding pre-training of GLIP allows for the effective detection of well-described distinct objects, its established structure struggles in distinguishing highly similar and densely packed within-class target nuclei. To alleviate the architectural constraints and prevent forgetting the generalizable knowledge of GLIP, we further design a self-trained knowledge distillation (SKD) framework, which iteratively optimizes the detection performance.

In our SKD framework, another advanced detection-specific network is employed as the final detector, and the knowledge distillation (KD) technique is adopted to assimilate GLIP’s knowledge \cite{hinton2015distilling}. Generally in KD, knowledge is transferred from the teacher network ${\mathcal{T}}$ to the student network ${\mathcal{S}}$ by minimizing the logit or feature difference \cite{hinton2015distilling, wang2021knowledge}. Without loss of generality, we choose YOLOX \cite{ge2021yolox}, a state-of-the-art detection network, as ${\mathcal{S}}$ in our method, aiming to improve detection performance. The modular design of YOLOX allows for easy switching of the feature extractor, which is kept identical to the image encoder of GLIP in our approach to facilitate weight initialization and knowledge distillation. Any segmentation architecture with this characteristic, including Mask-RCNN \cite{he2017mask}, can be employed as ${\mathcal{S}}$.

Due to the aim of achieving label-free detection, GLIP is adopted as the teacher network ${\mathcal{T}}$ with the initial predictions prompted through AttriPrompter serving as pseudo-labels. For simplicity, we denote the pseudo labels of the input image $x_{i}$ as $O_{i} =\{o_{1},\dots,o_{k_{i}}\}_{i}$, where $k_{i}$ represents the total number of pseudo labels. Denote the overall loss function of YOLOX as $L_{Det}(\cdot)$, the knowledge distillation loss as $L_{KD}$, with $n$ representing the number of training images, the objective of training ${\mathcal{S}}$ is formularized as:
\begin{equation}
\label{eq5}
L_{KD}=\left\|E_{\mathcal{V}}(x), E_{\mathcal{S}}(x)\right\|_1,
\end{equation}
\begin{equation}
\label{eq6}
L_{\mathcal{S}}=\frac{1}{n} \sum_{i=1}^n L_{Det}\left(O_i, \mathcal{H}\left(E_{\mathcal{S}}\left(x_i\right)\right)\right)+\alpha L_{KD}.
\end{equation}
The teacher is the GLIP image encoder $E_{\mathcal{V}}$, $E_{\mathcal{S}}$ and $\mathcal{H}$ represents the feature extractor and the segmentation head of YOLOX respectively, $\alpha$ is a weighting hyper-parameter. We follow Wang \textit{et al.} \cite{wang2023object} and use $L_{1}$ distance as the knowledge distillation loss $L_{KD}$.

In the context of SKD, to thoroughly distill the knowledge from GLIP, we further establish an iterative self-training mode \cite{dopido2013semisupervised}. This involves generating a new set of pseudo labels $O^{\prime}_{i} =\{o_{1},\dots,o_{k^{\prime}_{i}}\}_{i}$ with the converged $\mathcal{S}$ to initiate a new round of training, which enables the continuous refinement and polishing of the predicted boxes in a self-training manner:
\begin{equation}
L_{\mathcal{S}^{\prime}}=\frac{1}{n} \sum_{i=1}^n L_{Det}\left(O_i^{\prime}, \mathcal{S}^{\prime}\left(x_i\right)\right)+\alpha L_{KD},
\end{equation}
where $\mathcal{S}^{\prime}$ denotes the student in the new round. Since self-training is based on the EM optimization algorithm \cite{cheplygina2019not}, it propels our system to consistently enhance the prediction to achieve a better optimum.

\section{Experiments}
\label{experiments}
\subsection{Datasets}
\subsubsection{MoNuSeg} 
The MoNuSeg dataset consists of 30 nuclei images of size $1000\times1000$ with 658 nuclei per image on average \cite{kumar2017dataset}. We follow Kumar \textit{et al.} \cite{kumar2017dataset} and use 16 images for training and the rest 14 for testing. Four images are randomly chosen for validation at each training epoch. We extract 16 overlapped image patches of size $256\times256$ from each image and randomly crop them to $224\times224$ as the system input. It is worth noticing that in our work, training data is utilized for automatic prompting and producing pseudo labels for self-trained knowledge distillation while testing data and its ground truth are only utilized for evaluation.
\subsubsection{CoNSeP}
CoNSeP is a dataset for Colorectal Nuclear Segmentation and Phenotypes, consisting of 41 H\&E stained images each of size $1000\times1000$ extracted from 16 colorectal adenocarcinoma (CRA) WSIs \cite{graham2019hover}. For this dataset, we keep the same pre-processing procedure as MoNuSeg if not mentioned specifically. We use the default dataset partition: 26 images for training, 1 for validation, and the rest 14 for testing \cite{graham2019hover}.

\subsection{Evaluation Metrics and Implementation}
For evaluation, following COCO \cite{lin2014microsoft}, mAP, AP50, AP75, and AR are chosen as metrics.

In terms of experimental settings, we adopted the following configuration if not specified. We used 4 Nvidia RTX 3090 GPUs each with 24 GB of memory. For AttriPrompter, the VQA weights of BLIP finetuned on ViT-B and CapFilt-L  \cite{li2022blip} were adopted to generate [shape] and [color] attributes. The VQA question is: ``what is the color/shape of the nuclei''. We employed the pre-trained interface of GPT \cite{radford2019language}, specifically ChatGPT 3.5, for attribute augmentation with inquiries as: ``give the synonyms of \textunderscore\textunderscore, which can be used to describe the nuclei in a H\&E stained pathological image'' and ``give the words describing different degrees of \textunderscore\textunderscore''. For the relevance sorting, we chose 9 as $N$ to select prompts with high relevance. The weights used for GLIP is GLIP-L. As for self-trained knowledge distillation, we modify the feature extractor of YOLOX to be identical to the image encoder of GLIP and initialize with pre-trained weights. The other training settings of YOLOX are kept default \cite{ge2021yolox}. As for the self-training mode, we followed the standard self-training methodology described in \cite{dopido2013semisupervised}. The weighting hyper-parameter $\alpha$ for distillation loss is chosen as 1 after 4-fold cross-validation on MoNuSeg. The maximum of pseudo seeds $k_i$ is set as 20 per image if not specifically mentioned.

\begin{table}[t]
\caption{Comparison results on MoNuSeg \cite{kumar2017dataset} and CoNSeP \cite{graham2019hover}. Methods with a ``*'' mark are based on medical images, and those with a ``**'' mark are specifically based on H\&E images. The rest are originally based on natural images and were implemented for nuclei detection.} The best results of unsupervised methods are marked in \textbf{bold}.
\begin{center}
\resizebox{\columnwidth}{!}{
\begin{tabular}{ccccccccc}
\toprule[1.5pt]
\multirow{2}{*}{\textbf{Method}} & \multicolumn{4}{c}{\textbf{MoNuSeg}} & \multicolumn{4}{c}{\textbf{CoNSeP}} \\ \cmidrule(r){2-5} \cmidrule(r){6-9} 
 & \textbf{mAP} & \textbf{AP50} & \textbf{AP75} & \textbf{AR} & \textbf{mAP} & \textbf{AP50} & \textbf{AP75} & \textbf{AR} \\ \midrule[1.2pt]
\multicolumn{9}{c}{\textbf{Fully-supervised baseline}} \\
\textbf{YOLOX} & 0.447 & 0.832 & 0.437 & 0.528 & 0.388 & 0.726 & 0.384 & 0.495 \\ \hline
\multicolumn{9}{c}{\textbf{Weakly-supervised comparison}} \\
\textbf{WSPointA(2019)**\cite{qu2019weakly}} & 0.341 & 0.683 & 0.272 & 0.423 & 0.301 & 0.593 & 0.277 & 0.362 \\
\textbf{WSPPointA(2020)**\cite{qu2020weakly}} & 0.367 & 0.763 & 0.302 & 0.453 & 0.308 & 0.607 & 0.281 & 0.373 \\
\textbf{WSMixedA(2020)**\cite{qu2020nuclei}} & 0.374 & 0.771 & 0.374 & 0.488 & 0.330 & 0.653 & 0.307 & 0.412 \\
\textbf{WNSeg(2022)**\cite{liu2022weakly}} & 0.404 & 0.797 & 0.416 & 0.505 & 0.364 & 0.675 & 0.329 & 0.434 \\ \hline
\multicolumn{9}{c}{\textbf{Unsupervised comparison: Non-VLPM-based}} \\
\textbf{SSNS(2020)**\cite{sahasrabudhe2020self}} & 0.354 & 0.739 & 0.288 & 0.441 & 0.243 & 0.458 & 0.215 & 0.294 \\
\textbf{PDAM(2020)*\cite{liu2020pdam}} & 0.275 & 0.588 & 0.228 & 0.380 & 0.166 & 0.364 & 0.132 & 0.298 \\
\textbf{DARCNN(2021)*\cite{hsu2021darcnn}} & 0.262 & 0.476 & 0.267 & 0.356 & 0.173 & 0.388 & 0.129 & 0.310 \\
\textbf{Freesolo(2022)\cite{wang2022freesolo}} & 0.292 & 0.607 & 0.248 & 0.393 & 0.222 & 0.393 & 0.246 & 0.370 \\
\textbf{SOP(2022)*\cite{le2022unsupervised}} & 0.235 & 0.609 & 0.096 & 0.351 & 0.171 & 0.387 & 0.127 & 0.304 \\
\textbf{PSM(2022)**\cite{chen2022unsupervised}} & 0.227 & 0.447 & 0.218 & 0.375 & 0.142 & 0.323 & 0.102 & 0.286 \\
\textbf{CutLER(2023)\cite{wang2023cut}} & 0.115 & 0.213 & 0.117 & 0.184 & 0.089 & 0.181 & 0.049 & 0.155 \\ \hline
\multicolumn{9}{c}{\textbf{Unsupervised comparison: VLPM-based}} \\
\textbf{VL-PLM(2022)\cite{zhao2022exploiting}} & 0.333 & 0.582 & 0.342 & 0.501 & 0.159 & 0.357 & 0.120 & 0.291 \\
\textbf{VLDet(2023)\cite{lin2022learning}} & 0.173 & 0.407 & 0.112 & 0.263 & 0.110 & 0.296 & 0.053 & 0.245 \\
\textbf{MIU-VL(2023)*\cite{qin2022medical}} & 0.070 & 0.118 & 0.076 & 0.126 & 0.066 & 0.133 & 0.045 & 0.121 \\
\textbf{VLPM-NuD(2023)**\cite{wu2023zeroshot}} & 0.416 & 0.808 & 0.382 & 0.502 & 0.359 & 0.668 & 0.363 & 0.480 \\
\textbf{Ours**} & \textbf{0.425} & \textbf{0.819} & \textbf{0.401} & \textbf{0.511} & \textbf{0.367} & \textbf{0.669} & \textbf{0.376} & \textbf{0.488} \\ \bottomrule[1.5pt]
\end{tabular}
}
\label{tab:comparison}
\end{center}
\end{table}

\begin{figure*}[t]
\includegraphics[width=\linewidth]{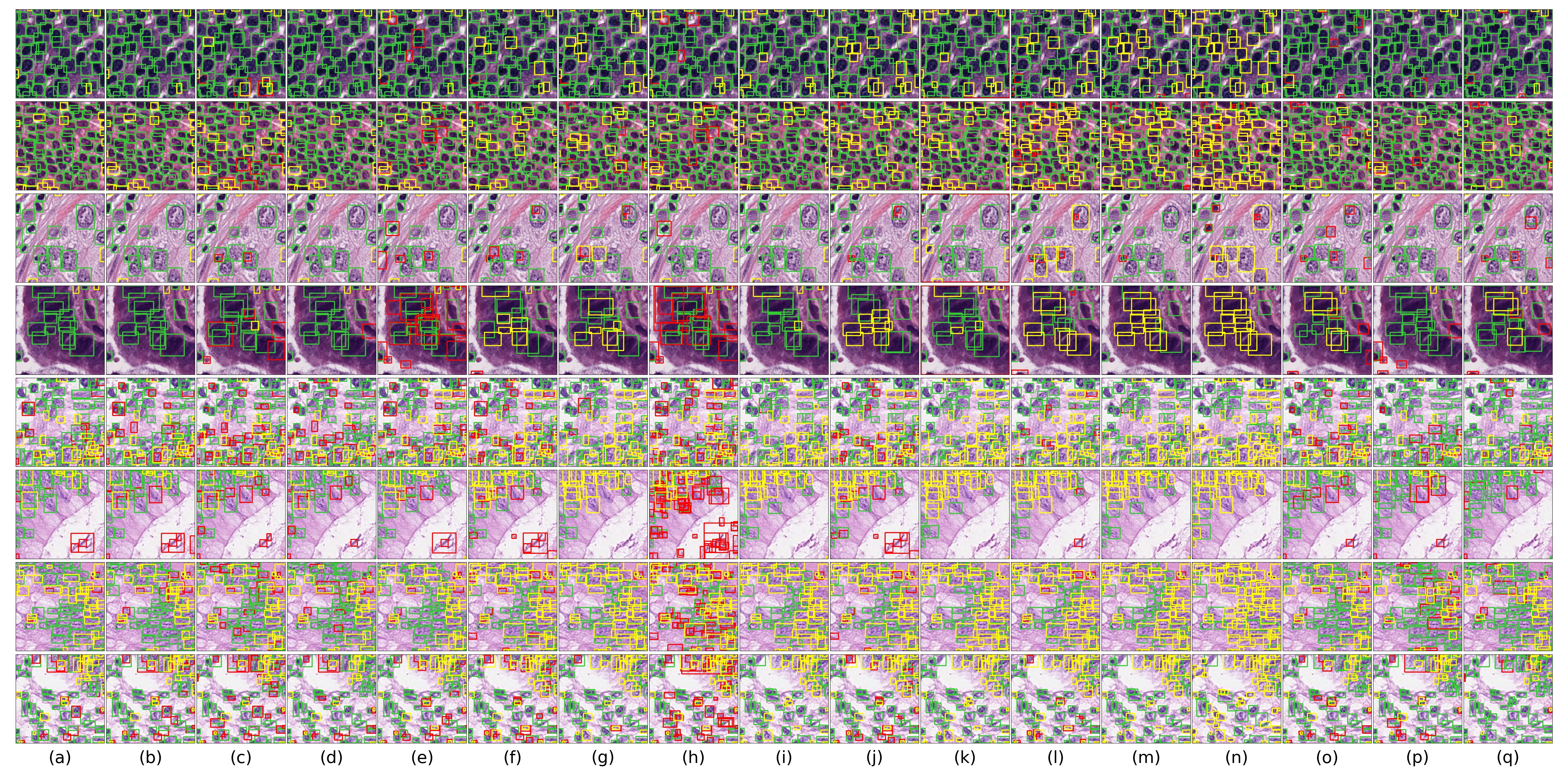}
\caption{Comparison output visualizations. The $1^{\text{st}} \sim 4^{\text{th}}$ rows: MoNuSeg, the $5^{\text{th}} \sim 8^{\text{th}}$ row: CoNSeP. Column: \textcolor{darkcyan}{(a)} WSPointA; \textcolor{darkcyan}{(b)} WSPPointA; \textcolor{darkcyan}{(c)} WSMixedA; \textcolor{darkcyan}{(d)} WNSeg; \textcolor{darkcyan}{(e)} SSNS; \textcolor{darkcyan}{(f)} PDAM; \textcolor{darkcyan}{(g)} DARCNN; \textcolor{darkcyan}{(h)} Freesolo; \textcolor{darkcyan}{(i)} SOP; \textcolor{darkcyan}{(j)} PSM; \textcolor{darkcyan}{(k)} CutLER; \textcolor{darkcyan}{(l)} VL-PLM; \textcolor{darkcyan}{(m)} VLDet; \textcolor{darkcyan}{(n)} MIU-VL; \textcolor{darkcyan}{(o)} VLPM-NuD; \textcolor{darkcyan}{(p)} ours; \textcolor{darkcyan}{(q)} fully-supervised YOLOX. Green, red, and yellow boxes denote true positives, false positives, and false negatives.} \label{fig:comparison}
\end{figure*}

\subsection{Comparison}

We compare our proposed method with its fully supervised counterparts and the current 11 state-of-the-art (SOTA) unsupervised object detection methods. Furthermore, for a thorough comparison, we also provide results from four weakly-supervised methods: WSPointA (2019) \cite{qu2019weakly}, WSPPointA (2020) \cite{qu2020weakly}, WSMixedA (2020) \cite{qu2020nuclei}, WNSeg (2022) \cite{liu2022weakly}. These four weakly-supervised methods use point annotations for supervision. We generate the point annotations for these point-based weakly-supervised methods by computing the central point of each nuclear mask following \cite{qu2019weakly}. Since the newly added weakly-supervised comparisons are all instance segmentation methods, we convert their predictions into detection results by using the bounding boxes of each predicted instance (expanded by two pixels). Unsupervised comparison methods include seven non-VLPM-based methods and four VLPM-based methods. The non-VLPM-based methods consist of SSNS (2020) \cite{sahasrabudhe2020self}, PDAM (2020) \cite{liu2020pdam}, DARCNN (2021) \cite{hsu2021darcnn}, Freesolo (2022) \cite{wang2022freesolo}, SOP (2022) \cite{le2022unsupervised}, PSM (2022) \cite{chen2022unsupervised}, and CutLER (2023) \cite{wang2023cut}. Most of these are specifically designed for biomedical microscopic imaging, except for Freesolo and CutLER. The VLPM-based methods include VL-PLM (2022) \cite{zhao2022exploiting}, VLDet (2023) \cite{lin2022learning}, MIU-VL (2023) \cite{qin2022medical}, and our previously published work VLPM-NuD (2023) \cite{wu2023zeroshot}. Note that MIU-VL and VLPM-NuD are both based on GLIP. We modify the key hyper-parameters of these methods to maximize their effectiveness on nuclei datasets. Specifically, for PDAM, DARCNN, CutLER, and VLDet, the maximum detection number is set to 100. For SSNS, the number of epochs is set to 1000. In Freesolo, the original self-training instance segmentation architecture is replaced with YOLOX to enhance detection outcomes, and the seed mask is simplified to a bounding box as the input for self-training. We keep the default values for other hyper-parameters or settings unless specifically mentioned.

\begin{table}[h]
\caption{Ablation study on automatic versus manual prompt design. This study excludes self-trained knowledge distillation for direct prompt quality evaluation. The best results are marked in \textbf{bold}.}
\label{tab:prompt}
\begin{center}
\resizebox{0.9\columnwidth}{!}{
\renewcommand{\arraystretch}{1.1}
\begin{tabular}{ccccccc}
\toprule[1.5pt]
\multicolumn{3}{c}{\multirow{2}{*}{\textbf{Prompt Design}}} & \multicolumn{4}{c}{\textbf{MoNuSeg}} \\ \cline{4-7} 
\multicolumn{3}{c}{} & \textbf{mAP} & \textbf{AP50} & \textbf{AP75} & \textbf{AR} \\ \midrule[1.2pt]
\multicolumn{3}{c}{\textbf{Manual}} & 0.151 & 0.259 & 0.166 & 0.210 \\ \hline
\multirow{10}{*}{\textbf{Auto}} & \multicolumn{2}{c}{\textbf{MIU-VL(2023)\cite{qin2022medical}}} & 0.070 & 0.118 & 0.076 & 0.126 \\ \cline{2-7} 
 & \textbf{Attribute} & \textbf{Noun} &  &  &  &  \\ \cline{2-7} 
 & \multirow{2}{*}{\textbf{\textbackslash{}}} & \textbf{'Nuclei.'} & 0.034 & 0.057 & 0.038 & 0.058 \\
 &  & \textbf{Aug. {[}noun.{]}} & 0.029 & 0.054 & 0.029 & 0.059 \\ \cline{2-7} 
 & \textbf{{[}shape{]}} & \textbf{``Nuclei.''} & 0.097 & 0.181 & 0.097 & 0.155 \\
 & \textbf{Aug. {[}shape{]}} & \textbf{``Nuclei.''} & 0.124 & 0.255 & 0.124 & 0.191 \\ \cline{2-7} 
 & \textbf{{[}color{]}} & \textbf{``Nuclei.''} & 0.009 & 0.016 & 0.010 & 0.024 \\
 & \textbf{Aug. {[}color{]}} & \textbf{``Nuclei.''} & 0.064 & 0.115 & 0.066 & 0.112 \\ \cline{2-7} 
 & \textbf{{[}shape{]}{[}color{]}} & \textbf{``Nuclei.''} & 0.155 & 0.285 & 0.152 & \textbf{0.239} \\
 & \textbf{Aug. {[}shape{]}{[}color{]}} & \textbf{``Nuclei.''} & \textbf{0.179} & \textbf{0.310} & \textbf{0.193} & 0.220 \\ \bottomrule[1.5pt]
\end{tabular}
}
\end{center}
\end{table}

Our method outperforms other unsupervised approaches in H\&E nuclei detection and even achieves comparable performance with the fully-supervised counterpart (see \autoref{tab:comparison}). Among non-VLPM-based methods, performance is generally poor, with SSNS, designed for histopathological images, showing slightly better results. However, open-vocabulary VLPM-based methods, such as VL-PLM and VLDet, perform mediocrely, as they are not designed for medical images. Astonishingly, GLIP-based MIU-VL, designed for medical images, performs the worst. This is because MIU-VL, being a prompt generation method, lacks adaptation for dense nuclei and is more suitable for sparse medical image detection tasks like polyp or tumor detection. In contrast, our method achieves favorable results, and with improvements like adding degree augmentation to auto-prompting and incorporating knowledge distillation, it attains optimal performance. Moreover, while weakly-supervised methods generally outperform other unsupervised comparison methods, our method performs even better without requiring manual annotations. \autoref{fig:comparison} visually presents the detection results, showcasing our approach's ability to exploit VLPMs' potential, mitigating missed detection and false positives. our method pushes the boundaries of unsupervised H\&E nuclei detection.

\subsection{Ablation Studies}
We conducted ablation studies on the key components of the proposed framework using the MoNuSeg dataset.
\begin{table}[t]
\caption{Ablation study on AttriPrompter components. This study excludes self-trained knowledge distillation for direct prompt quality evaluation. The best results are marked in \textbf{bold}.}
\label{tab:promptcom}
\begin{center}
\resizebox{0.9\columnwidth}{!}{
\begin{tabular}{ccccccc}
\toprule[1.5pt]
\textbf{\begin{tabular}[c]{@{}c@{}}Synonym \\ Augmentation\end{tabular}} &
  \textbf{\begin{tabular}[c]{@{}c@{}}Degree\\ Augmentation\end{tabular}} &
  \textbf{\begin{tabular}[c]{@{}c@{}}Relevance\\ Sorting\end{tabular}} &
  \textbf{mAP} &
  \textbf{AP50} &
  \textbf{AP75} &
  \textbf{AR} \\ \midrule[1.2pt]
  &   &   & 0.150           & 0.275          & 0.150           & 0.228          \\
\checkmark &   &   & 0.160           & 0.285          & 0.173          & 0.201          \\
  & \checkmark &   & 0.176          & 0.304          & 0.187          & 0.213          \\
  &   & \checkmark & 0.155          & 0.290           & 0.152          & \textbf{0.239} \\
\checkmark &   & \checkmark & 0.174          & 0.301          & 0.188          & 0.210           \\
  & \checkmark & \checkmark & 0.178          & \textbf{0.314} & 0.187          & 0.217          \\
\checkmark & \checkmark &   & 0.177          & 0.309          & \textbf{0.193} & 0.220           \\
\checkmark & \checkmark & \checkmark & \textbf{0.179} & 0.310           & \textbf{0.193} & 0.220           \\ \bottomrule[1.5pt]
\end{tabular}
}
\end{center}
\vskip -0.15in
\end{table}

\subsubsection{Automatic vs. Manual Prompt Design} The ablation study on automatic vs. manual prompt design for zero-shot detection is detailed in \autoref{tab:prompt}. In this study, self-trained knowledge distillation is omitted to directly evaluate prompt quality. Instead, we solely reported results for zero-shot detection using prompt sequences directly input into GLIP, maintaining consistent experimental settings unless specified. The first row shows the manual design, where we invited three pathologists to independently design prompts with nuclei descriptions for zero-shot detection, ultimately reporting the average of the results. An example of a prompt is ``Nuclei, characterized by their distinct chromatin patterns and nucleoli, typically stained dark blue or purple''. The second row presents the results of another automatic prompt generation method, MIU-VL.  Subsequent rows employ prompts automatically generated by VLPMs, combining objectives and nouns, such as ``circular purple nuclei''. Relevance sorting is conducted before prompts are fed into GLIP. \textbf{`Aug.'} indicates prompts that underwent GPT augmentation. Notably, augmentation extended to nouns (the 3rd row), with noun lists before and after augmentation being [``Nuclei''] and [``Nuclei'', ``Nucleus'', ``Cyteblast'', ``Karyon''], respectively. 

Results in \autoref{tab:prompt} indicate that our AttriPrompter's (the last row) performance can match or exceed manual design, which tends to be empirical and subject to subjective bias. As attribute descriptors evolve from solely addressing shape or color to encompassing both, GLIP's performance steadily improves. Notably, noun augmentation doesn't yield optimal results compared to using [``Nuclei''] alone, possibly because the augmented medical synonyms are uncommon and were absent in GLIP's pre-training data. In contrast, attribute word augmentation, involving common descriptors for natural scenes, is generally more effective. It is noteworthy that the comparative method, MIU-VL, performs poorly. This is due to its incorporation of the location attribute, which is counterproductive for dense nuclei detection.

\subsubsection{AttriPrompter Components} As detailed in \autoref{tab:promptcom}, each component contributes to the enhanced effectiveness of the prompts, with the best mAP achieved when combined. Notably, the method of employing only synonym augmentation (the $2^{nd}$ row) aligns with our approach in previously published work \cite{wu2023zeroshot}. Degree augmentation stands out for its ability to generate detailed terms with finer granularity, assisting GLIP in accurately distinguishing target nuclei from similar entities, like background or non-target tissues. This is also evidenced in \autoref{tab:prompt}, where prompts relying solely on non-augmented color descriptors perform poorly, likely due to the color similarity between the background and nuclei in H\&E images. However, the inclusion of degree augmentation markedly improves detection efficacy.


\begin{table}[t]
\caption{Results of adopting different pseudo label generation methods for self-training. The results labeled "Ours" did not incorporate knowledge distillation for fair comparison. The best results are marked in \textbf{bold}.}
\label{tab:selftraining}
\begin{center}
\resizebox{\columnwidth}{!}{
\begin{tabular}{ccccccccccccc}
\toprule[1.5pt]
\multirow{2}{*}{\textbf{Round}} & \multicolumn{4}{c}{\textbf{Superpixels}} & \multicolumn{4}{c}{\textbf{MIU-VL}} & \multicolumn{4}{c}{\textbf{AttriPrompter (Ours)}} \\ \cmidrule(r){2-5} \cmidrule(r){6-9} \cmidrule(r){10-13}  
 & \textbf{mAP} & \textbf{AP50} & \textbf{AP75} & \textbf{AR} & \textbf{mAP} & \textbf{AP50} & \textbf{AP75} & \textbf{AR} & \textbf{mAP} & \textbf{AP50} & \textbf{AP75} & \textbf{AR} \\ \midrule[1.2pt]
\textbf{Round 0} & 0.027 & 0.075 & 0.012 & 0.035 & 0.070 & 0.118 & 0.076 & 0.126 & 0.179 & 0.310 & 0.193 & 0.220 \\
\textbf{Round 1} & 0.260 & 0.612 & 0.162 & 0.373 & 0.221 & 0.486 & 0.208 & 0.370 & 0.302 & 0.693 & 0.248 & 0.393 \\
\textbf{Round 2} & 0.272 & 0.617 & 0.183 & 0.392 & 0.242 & 0.495 & 0.239 & 0.381 & 0.358 & 0.724 & 0.326 & 0.430 \\
\textbf{Round 3} & 0.284 & 0.655 & 0.169 & 0.404 & 0.266 & 0.511 & 0.265 & 0.409 & \textbf{0.363} & 0.730 & 0.327 & 0.433 \\
\textbf{Round 4} & 0.279 & 0.614 & 0.213 & 0.389 & 0.269 & 0.515 & 0.266 & 0.416 & 0.361 & \textbf{0.732} & \textbf{0.331} & \textbf{0.435} \\ \bottomrule[1.5pt]
\end{tabular}
}
\end{center}
\vskip -0.15in
\end{table}

\subsubsection{Adopting Different Pseudo Label Generation Methods for Self-training} Self-trained knowledge distillation facilitates continuous optimization of an advanced detection-specific network, concurrently preserving GLIP's superior object-level knowledge, thereby optimizing the zero-shot nuclei detection potential. However, the primary factor enabling our method to rival fully supervised approaches is the efficacy of AttriPrompter. To elucidate this point, we outline the optimization process in \autoref{tab:selftraining}. We employed a standard unsupervised detection technique, Superpixels \cite{achanta2012slic}, and an automated GLIP prompt generation method for medical imaging, MIU-VL \cite{qin2022medical}, to generate zero-shot pseudo labels. These pseudo labels, derived from Superpixels or MIU-VL, were input into a self-training framework modeled on the YOLOX segmentation architecture. This framework was consistent with our approach, except in the aspects of pseudo label generation and knowledge distillation. For a fair comparison, the results labeled ``AttriPrompter'' in \autoref{tab:selftraining} did not incorporate knowledge distillation. Within the table, ``Round 0'' indicates a direct pseudo label to ground truth comparison, ``Round 1'' reflects the initial convergence of the student network trained on pseudo labels, and subsequent rounds involve using the previous round's converged student network predictions as new pseudo labels. Results are suboptimal when Superpixels or MIU-VL are used for seed pseudo label generation. This outcome underscores that the high performance of our method primarily hinges on the proficient harnessing of knowledge from VLPMs and the effectiveness of AttriPrompter, rather than the detection architecture or the self-training process. 

\subsubsection{Pseudo labels vs. Real labels} To evaluate the pseudo labels more clearly, we compared them directly with real labels on the testing sets of MoNuSeg and CoNSeP. Besides mAP, AP50, AP75, and AR, we introduced the mean maximum IoU between each pseudo-label and all real labels, denoted as mIoU$_{nearest}$. This metric represents the average discrepancy between the predicted pseudo-label and the nearest real label. The results are presented in \autoref{tab:pseudo-real}. Due to the high density of nuclei in H\&E images, initiatory GLIP-generated pseudo labels exhibit numerous omissions, resulting in low Recall and consequently low AP (i.e. the area under the Precision-Recall curve) and AR (i.e. Average Recall). However, the mIoU$_{nearest}$ value is relatively high, suggesting that the pseudo labels have promising accuracy.

\begin{table}[t]
\caption{Pseudo labels quality: direct comparison with real labels.}
\label{tab:pseudo-real}
\begin{center}
\resizebox{0.86\columnwidth}{!}{
\begin{tabular}{cccccc}
\toprule[1.5pt]
\textbf{Dataset} & \textbf{mAP} & \textbf{AP50} & \textbf{AP75} & \textbf{AR} & \textbf{mIoU$_{nearest}$} \\ \midrule[1.2pt]
\textbf{MoNuSeg} & 0.178 & 0.312 & 0.195 & 0.217 & 0.742 \\
\textbf{CoNSeP} & 0.096 & 0.178 & 0.101 & 0.146 & 0.690 \\ \bottomrule[1.5pt]
\end{tabular}
}
\end{center}
\vskip -0.1in
\end{table}

\begin{table}[t]
\caption{Pseudo labels quality: Comparison between using pseudo labels generated by Attriprompter and using different proportions of real labels in the SKD framework. The best and second-best values are highlighted in \textbf{bold} and {\ul underlined}, respectively.}
\label{tab:semi-supervised}
\begin{center}
\resizebox{0.78\columnwidth}{!}{
\begin{tabular}{cccccc}
\toprule[1.5pt]
\multicolumn{2}{c}{\textbf{Labels}} & \textbf{mAP} & \textbf{AP50} & \textbf{AP75} & \textbf{AR} \\ \midrule[1.2pt]
\multirow{4}{*}{\textbf{\begin{tabular}[c]{@{}c@{}}Real\\ Labels\end{tabular}}} & \textbf{1\%} & 0.185 & 0.588 & 0.037 & 0.276 \\
 & \textbf{5\%} & 0.345 & 0.728 & 0.279 & 0.429 \\
 & \textbf{10\%} & 0.378 & 0.761 & 0.332 & 0.458 \\
 & \textbf{50\%} & \textbf{0.443} & \textbf{0.819} & \textbf{0.448} & \textbf{0.516} \\ \midrule
\multicolumn{2}{c}{\textbf{Pseudo Labels}} & {\ul 0.425} & \textbf{0.819} & {\ul 0.401} & {\ul 0.511} \\ \bottomrule[1.5pt]
\end{tabular}
}
\end{center}
\vskip -0.15in
\end{table}

Additionally, we conducted experiments in a semi-supervised learning setting using different proportions of real labeled data in our SKD framework on MoNuSeg to further demonstrate the quality of the generated pseudo-labels. The results are listed in \autoref{tab:semi-supervised}. The percentages in the table represent instance-level proportions, e.g., 50\% means that ground-truth boxes for 50\% of instances in each training image are used as supervision. As shown in the table, the generated pseudo-labels outperform the use of 10\% real labels. Although they do not surpass the performance of using 50\% real labels, our pseudo label generation process is entirely automated, significantly reducing the annotation burden.

\subsubsection{Analysis of Optimal Prompts} The generation and augmentation of automatic prompts are the main points of focus of our work. Therefore, we provided a quantitative analysis of the optimal prompts on the MoNuSeg dataset in \autoref{tab:detailPrompt}. Specifically, we included prompts sorted by Relevance Sorting, their relevance, and the mAP values predicted by GLIP for each individual prompt. The top N=9 prompts were selected as optimal prompts after cross-validation, while the remaining prompts with lower relevance were discarded; a few of these less relevant prompts were provided as comparison samples for simplicity.

\begin{table}[t]
\caption{Quantitative analysis of the optimal prompts on the MoNuSeg dataset. The top N=9 prompts are selected after cross-validation. The remaining prompts are discarded and marked with a gray background.}
\label{tab:detailPrompt}
\begin{center}
\resizebox{0.98\columnwidth}{!}{
\begin{tabular}{cccc}
\toprule[1.5pt]
\textbf{Order} & \textbf{Prompt} & \textbf{Relevance} & \textbf{mAP} \\ \midrule[1.2pt]
1 & ``slightly round dark purple nuclei.'' & 0.317 & 0.116 \\
2 & ``slightly round purple nuclei.'' & 0.315 & 0.128 \\
3 & ``moderately round purple nuclei.'' & 0.312 & 0.104 \\
4 & ``moderately round dark purple nuclei.'' & 0.311 & 0.103 \\
5 & ``slightly round violet nuclei.'' & 0.311 & 0.099 \\
6 & ``circular purple nuclei.'' & 0.306 & 0.063 \\
7 & ``circular dark purple nuclei.'' & 0.304 & 0.067 \\
8 & ``circular violet nuclei.'' & 0.303 & 0.061 \\
9 & ``moderately round violet nuclei.'' & 0.302 & 0.057 \\
\rowcolor{lightgray} 
10 & ``mostly round rich purple nuclei.'' & 0.299 & 0.054 \\
\rowcolor{lightgray} 
11 & ``moderately round rich purple nuclei.'' & 0.299 & 0.051 \\
\rowcolor{lightgray} 
12 & ``elliptical purple-blue nuclei.'' & 0.298 & 0.041 \\
\rowcolor{lightgray} 
13 & ``elongated dark purple nuclei.'' & 0.296 & 0.040 \\
\rowcolor{lightgray} 
14 & ``slightly circular plum-colored nuclei.'' & 0.294 & 0.033 \\
\rowcolor{lightgray} 
15 & ``elliptical vibrant purple nuclei.'' & 0.294 & 0.032 \\
\rowcolor{lightgray} 
\multicolumn{4}{c}{\cellcolor{lightgray}...} \\ \bottomrule[1.5pt]
\end{tabular}
}
\end{center}
\vskip -0.15in
\end{table}

From \autoref{tab:detailPrompt}, we observe that prompts containing ``slightly round'' and ``dark purple'' yield better prediction results. This can be interpreted as these descriptions using common words that align well with the natural text used during GLIP pre-training, allowing for effective transfer of GLIP's pre-trained knowledge and matching the basic characteristics of nuclei in H\&E images \cite{shi2022nuclei}. This observation aligns with our findings in \autoref{sec:promptana}. Other effective words describing shape, such as ``moderately round'' and ``circular,'' and color, such as ``purple'' and ``violet,'' also correspond well to the characteristics of stained nuclei. In contrast, less effective words like ``elliptical'' and ``elongated'' were not found in the top 5000 words of the Corpus of Contemporary American English \footnote{\href{https://www.english-corpora.org/coca/}{https://www.english-corpora.org/coca/}. The word frequency list is provided in WordFrequency.info: \href{https://www.wordfrequency.info/}{https://www.wordfrequency.info/}}. Their relatively low frequency in natural language likely contributed to poorer performance in GLIP predictions.

\subsubsection{Knowledge Distillation Loss} The ablation study on the knowledge distillation loss $L_{KD}$ is detailed in \autoref{tab:KD}. To confirm its efficacy in preserving GLIP's prior knowledge, we ablated $L_{KD}$ under various prompt settings, maintaining consistent experimental configurations with previous sections. Results show that, irrespective of the prompt, $L_{KD}$ more effectively transfers object-level knowledge from GLIP to the detection network, ultimately boosting the efficacy of self-training. When coupled with AttriPrompter, self-trained knowledge distillation maximizes the overall performance of the system in zero-shot nuclei detection.

\begin{table}[t]
\caption{Ablation study on the knowledge distillation loss $L_{KD}$. Not using $L_{KD}$ implies reliance solely on $L_{Det}$ for self-training. The best results are marked in \textbf{bold}.}
\label{tab:KD}
\begin{center}
\resizebox{0.9\columnwidth}{!}{
\begin{tabular}{cccccc}
\toprule[1.5pt]
\textbf{Prompt Design}                                          & \textbf{$L_{KD}$} & \textbf{mAP} & \textbf{AP50} & \textbf{AP75} & \textbf{AR} \\ \midrule[1.2pt]
\multirow{2}{*}{\textbf{Manual}}      &    & 0.409 & 0.801 & 0.383 & 0.493 \\
                                      & \checkmark & 0.418 & 0.811 & 0.392 & 0.504 \\ \hline
\multirow{2}{*}{\textbf{``Nuclei.''}} &    & 0.324 & 0.696 & 0.316 & 0.381 \\
                                      & \checkmark & 0.352 & 0.711 & 0.334 & 0.414 \\ \hline
\multirow{2}{*}{\textbf{AttriPrompter}} &          & 0.412        & 0.805         & 0.385         & 0.498       \\
                                      & \checkmark & \textbf{0.425} & \textbf{0.819} & \textbf{0.401} & \textbf{0.511} \\ \bottomrule[1.5pt]
\end{tabular}
}
\end{center}
\end{table}

\begin{table}[t]
\caption{Ablation study on the architecture of $\mathcal{S}$.}
\label{tab:architecture}
\begin{center}
\resizebox{0.9\columnwidth}{!}{
\begin{tabular}{cccccc}
\toprule[1.5pt]
\textbf{$\mathcal{S}$}                   & \textbf{Method}           & \textbf{mAP} & \textbf{AP50} & \textbf{AP75} & \textbf{AR} \\ \midrule[1.2pt]
\multirow{3}{*}{\textbf{YOLOX}}     & \textbf{Fully-supervised} & 0.447        & 0.832         & 0.437         & 0.528       \\
 & \textbf{VLPM-NuD} & 0.416 & 0.808 & 0.382 & 0.502 \\
 & \textbf{Ours}     & 0.425 & 0.819 & 0.401 & 0.511 \\ \hline
\multirow{3}{*}{\textbf{Mask-RCNN}} & \textbf{Fully-supervised} & 0.385        & 0.751         & 0.358         & 0.466       \\
 & \textbf{VLPM-NuD} & 0.350  & 0.718 & 0.321 & 0.412 \\
 & \textbf{Ours}     & 0.371 & 0.734 & 0.336 & 0.446 \\ \bottomrule[1.5pt]
\end{tabular}
}
\end{center}
\vskip -0.15in
\end{table}

\subsubsection{Different Student Detection Architecture}
Our proposed framework can achieve excellent results when using different detection architectures as the student network $\mathcal{S}$ for self-trained knowledge distillation. As shown in \autoref{tab:architecture}, we conducted experiments using Mask-RCNN \cite{he2017mask} as $\mathcal{S}$, employing the same self-trained knowledge distillation settings that were utilized when YOLOX served as $\mathcal{S}$. For the setting of Mask-RCNN, we set the maximum number of ground truth objects to 100 in one image, the non-max suppression threshold of RPN to 0.9 during training, and 0.7 during testing. As for other Mask-RCNN hyper-parameters, we keep the default setting used on COCO by He \textit{et al.} \cite{he2017mask}. Notably, the $2^{nd}$ and the $5^{th}$ rows show results of our previously published work \cite{wu2023zeroshot}. It can be observed that our framework also achieves comparable performance with fully supervised methods with Mask-RCNN. This illustrates the generality of the proposed method.

\subsection{Domain Shift Study}

\begin{figure}[t]
\includegraphics[width=\linewidth]{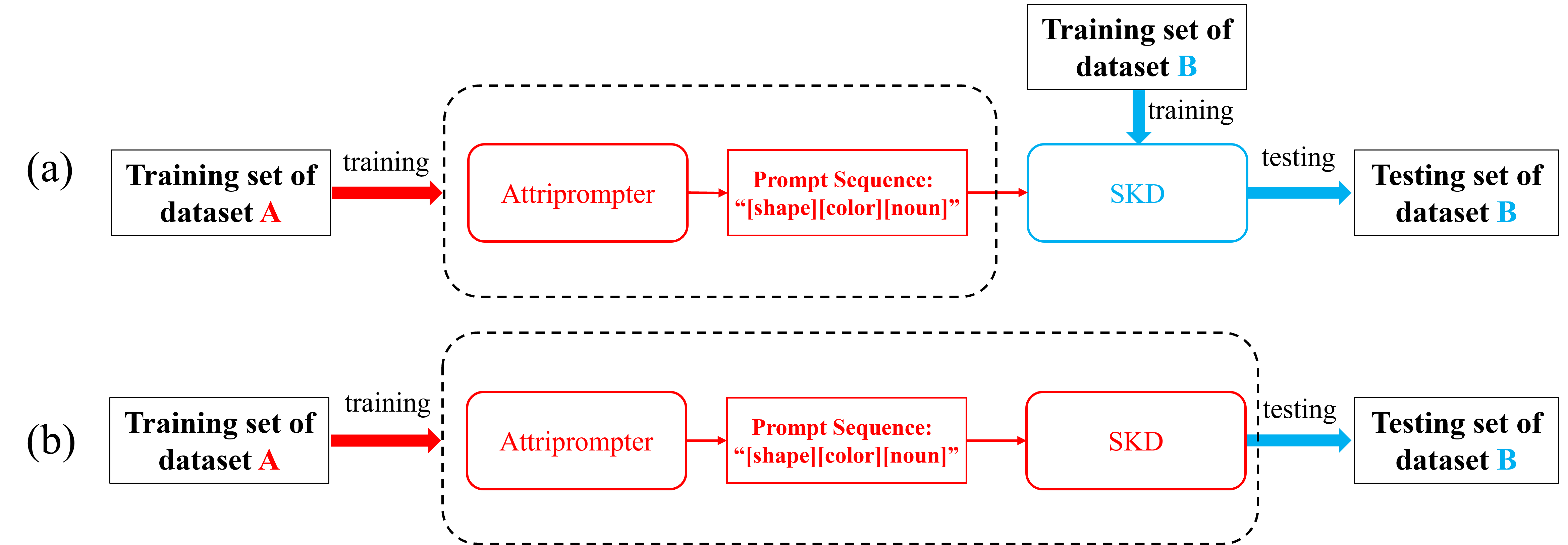}
\caption{Two cross-domain experiments: \textcolor{darkcyan}{(a)} the cross-domain capacity of Attriprompter, \textcolor{darkcyan}{(b)} the domain shift study for the whole system. ``Dataset A'' denotes the source domain, ``Dataset B'' denotes the target domain.} \label{fig:cross domain}
\end{figure}

One of the major challenges in medical imaging is related to domain shift, such as when images are captured in different medical centers or using different types of stains in histopathology images. Therefore, it is necessary to explore the performance of our system under domain shift. We conducted two different cross-domain experiments, illustrated in \autoref{fig:cross domain}.

\begin{table}[t]
\caption{Validation of the cross-domain capacity of Attriprompter.}
\label{tab:cross-attri}
\begin{center}
\resizebox{0.9\columnwidth}{!}{
\begin{tabular}{cccccc}
\toprule[1.5pt]
\textbf{\begin{tabular}[c]{@{}c@{}}Dataset used for\\      Attriprompter\end{tabular}} & \textbf{\begin{tabular}[c]{@{}c@{}}Dataset used for\\      SKD and Testing\end{tabular}} & \textbf{mAP} & \textbf{AP50} & \textbf{AP75} & \textbf{AR} \\ \midrule[1.2pt]
\textbf{MoNuSeg} & \textbf{MoNuSeg} & 0.425 & 0.819 & 0.401 & 0.511 \\
\textbf{CoNSeP} & \textbf{MoNuSeg} & 0.405 & 0.759 & 0.400 & 0.480 \\ \midrule
\textbf{CoNSeP} & \textbf{CoNSeP} & 0.367 & 0.669 & 0.376 & 0.488 \\
\textbf{MoNuSeg} & \textbf{CoNSeP} & 0.339 & 0.669 & 0.324 & 0.441 \\ \bottomrule[1.5pt]
\end{tabular}
}
\end{center}
\vskip -0.15in
\end{table}

The first experiment is a validation of the cross-domain capacity of Attriprompter. We used MoNuSeg or CoNSeP to generate prompts through Attriprompter and then used these prompts for self-trained knowledge distillation (SKD) and testing on the other dataset. The results are listed in \autoref{tab:cross-attri}. We found that using different datasets for Attriprompter and SKD results in a slight performance drop compared to using the same dataset. This is reasonable because MoNuSeg and CoNSeP may have small differences in nucleus color or shape, and the prompts generated by Attriprompter may not perfectly match the other dataset, yielding suboptimal pseudo labels for SKD and reducing performance. This indicates that our system performs best when the data used by Attriprompter closely matches the target data.

The second experiment tests the entire system, including SKD, under domain shift. Unlike the first experiment, this setting trains the whole system using source domain data and tests it on a different target domain (\autoref{fig:cross domain}\textcolor{darkcyan}{.\;(b)}), whereas the first experiment uses the target domain's training set for SKD (\autoref{fig:cross domain}\textcolor{darkcyan}{.\;(a)}). We compared standard zero-shot strategies (i.e., manual design), Attriprompter, and Attriprompter with SKD using different source and target datasets, i.e., training the whole system on MoNuSeg or CoNSeP and testing on the other. Additionally, we compared our method with the top three unsupervised methods from \autoref{tab:comparison}: SSNS, VL-PLM, and VLPM-NuD. The results are in \autoref{tab:domain shift}. In the table header, ``→'' indicates the direction from the source domain to the target domain, e.g., ``MoNuSeg → CoNSeP'' means training on MoNuSeg and testing on CoNSeP. The first row shows results from standard zero-shot strategies obtained by manually designing prompts for the target domain without any cross-domain process, marked with a gray background. ``Attriprompter'' results are from evaluating prompts generated in the source domain on the target domain.

Comparing \autoref{tab:comparison} and \autoref{tab:domain shift}, we observed a performance drop in our framework when facing domain shift. However, our method still outperforms the comparison methods under domain shift, achieving the best results. Notably, standard zero-shot strategies do not surpass Attriprompter and Attriprompter with SKD in cross-domain scenarios. This might be because nuclei in H\&E images from different domains have similarities to some extent, and Attriprompter uses common attribute words during prompt generation. Therefore, the prompts generated by Attriprompter perform barely well even in cross-domain scenarios. In contrast, standard zero-shot strategies introduce descriptions not covered by GLIP's semantic space, resulting in less effective performance compared to the cross-domain prompts generated by Attriprompter.

\section{Conclusion}

\begin{table}[t]
\caption{Domain shift study on the proposed framework. "→" indicates the direction from the source domain to the target domain. Standard zero-shot strategies do not undergo any cross-domain process and are marked with a gray background. The best results are marked in \textbf{bold}.}
\label{tab:domain shift}
\begin{center}
\resizebox{0.96\columnwidth}{!}{
\begin{tabular}{ccccccccc}
\toprule[1.5pt]
 & \multicolumn{4}{c}{\textbf{MoNuSeg → CoNSeP}} & \multicolumn{4}{c}{\textbf{CoNSeP → MoNuSeg}} \\ \cmidrule(r){2-5} \cmidrule(r){6-9}
\multirow{-2}{*}{\textbf{Method}} & \textbf{mAP} & \textbf{AP50} & \textbf{AP75} & \textbf{AR} & \textbf{mAP} & \textbf{AP50} & \textbf{AP75} & \textbf{AR} \\ \hline
\rowcolor{lightgray} 
\textbf{Standard Zero-shot (Manual)} & 0.062 & 0.131 & 0.055 & 0.111 & 0.151 & 0.259 & 0.166 & 0.210 \\
\textbf{SSNS (2020)\cite{sahasrabudhe2020self}} & 0.122 & 0.310 & 0.101 & 0.274 & 0.111 & 0.204 & 0.115 & 0.169 \\
\textbf{VL-PLM (2022)\cite{zhao2022exploiting}} & 0.118 & 0.307 & 0.097 & 0.275 & 0.096 & 0.175 & 0.108 & 0.133 \\
\textbf{VLPM-NuD(2023)\cite{wu2023zeroshot}} & 0.152 & 0.389 & 0.080 & 0.257 & 0.173 & 0.250 & 0.171 & 0.225 \\
\textbf{Attriprompter} & 0.082 & 0.182 & 0.061 & 0.124 & 0.168 & 0.287 & 0.175 & 0.199 \\
\textbf{Attriprompter + SKD} & \textbf{0.178} & \textbf{0.439} & \textbf{0.112} & \textbf{0.285} & \textbf{0.260} & \textbf{0.594} & \textbf{0.178} & \textbf{0.356} \\ \bottomrule[1.5pt]
\end{tabular}
}
\end{center}
\end{table}

This paper proposes a novel label-free nuclei detection framework leveraging VLPMs for the first time. Outperforming all existing unsupervised methods, our approach demonstrates exceptional generality. We propose an innovative auto-prompting pipeline, i.e. AttriPrompter, overcoming limitations of prior work, such as inaccurate descriptions and oversight in prompt sequence order. To address high nuclei density, a self-trained knowledge distillation framework is developed, reducing missed detections and false positives. The prompt sequence enables VLPMs to localize unfamiliar nuclei without prior medical image knowledge and serves as a paradigm for zero-shot transfer in other medical image modalities.

Given that this work is a pioneering exploration of using VLPMs for zero-shot nuclei detection, several limitations are inevitable: (1) Attriprompter currently generates descriptions only related to [shape] and [color], potentially overlooking other types of semantics. (2) The training overhead for SKD is substantial. (3) Performance drops when facing domain shift, indicating room for improvement. Future work will consider: (1) Exploring the automatic generation of a wider range of attributes. (2) Seeking efficient training methods to reduce overhead. (3) Integrating domain generalization training strategies to enhance cross-domain generalization. Additionally, future work will explore VLPMs' potential in more complex scenarios, including zero-shot nuclei grading and segmentation.



\normalem
\bibliographystyle{IEEEtran}
\bibliography{tmi.bib}

\begin{thebibliography}{10}
\providecommand{\url}[1]{#1}
\csname url@samestyle\endcsname
\providecommand{\newblock}{\relax}
\providecommand{\bibinfo}[2]{#2}
\providecommand{\BIBentrySTDinterwordspacing}{\spaceskip=0pt\relax}
\providecommand{\BIBentryALTinterwordstretchfactor}{4}
\providecommand{\BIBentryALTinterwordspacing}{\spaceskip=\fontdimen2\font plus
\BIBentryALTinterwordstretchfactor\fontdimen3\font minus \fontdimen4\font\relax}
\providecommand{\BIBforeignlanguage}[2]{{%
\expandafter\ifx\csname l@#1\endcsname\relax
\typeout{** WARNING: IEEEtran.bst: No hyphenation pattern has been}%
\typeout{** loaded for the language `#1'. Using the pattern for}%
\typeout{** the default language instead.}%
\else
\language=\csname l@#1\endcsname
\fi
#2}}
\providecommand{\BIBdecl}{\relax}
\BIBdecl

\bibitem{gleason1992histologic}
D.~F. Gleason, ``Histologic grading of prostate cancer: a perspective,'' \emph{Human pathology}, vol.~23, no.~3, pp. 273--279, 1992.

\bibitem{mahanta2021ihc}
L.~B. Mahanta, E.~Hussain, N.~Das, L.~Kakoti, and M.~Chowdhury, ``Ihc-net: A fully convolutional neural network for automated nuclear segmentation and ensemble classification for allred scoring in breast pathology,'' \emph{Applied Soft Computing}, vol. 103, p. 107136, 2021.

\bibitem{graham2019hover}
S.~Graham, Q.~D. Vu, S.~E.~A. Raza, A.~Azam, Y.~W. Tsang, J.~T. Kwak, and N.~Rajpoot, ``Hover-net: Simultaneous segmentation and classification of nuclei in multi-tissue histology images,'' \emph{Medical Image Analysis}, vol.~58, p. 101563, 2019.

\bibitem{yi2019multi}
J.~Yi, P.~Wu, Q.~Huang, H.~Qu, B.~Liu, D.~J. Hoeppner, and D.~N. Metaxas, ``Multi-scale cell instance segmentation with keypoint graph based bounding boxes,'' in \emph{International Conference on Medical Image Computing and Computer-Assisted Intervention (MICCAI)}.\hskip 1em plus 0.5em minus 0.4em\relax Springer, 2019, pp. 369--377.

\bibitem{zhou2023cyclic}
Y.~Zhou, Y.~Wu, Z.~Wang, B.~Wei, M.~Lai, J.~Shou, Y.~Fan, and Y.~Xu, ``Cyclic learning: Bridging image-level labels and nuclei instance segmentation,'' \emph{IEEE Transactions on Medical Imaging}, vol.~42, no.~10, pp. 3104--3116, 2023.

\bibitem{mouelhi2018fast}
A.~Mouelhi, H.~Rmili, J.~B. Ali, M.~Sayadi, R.~Doghri, and K.~Mrad, ``Fast unsupervised nuclear segmentation and classification scheme for automatic allred cancer scoring in immunohistochemical breast tissue images,'' \emph{Computer methods and programs in biomedicine}, vol. 165, pp. 37--51, 2018.

\bibitem{sahasrabudhe2020self}
M.~Sahasrabudhe, S.~Christodoulidis, R.~Salgado, S.~Michiels, S.~Loi, F.~Andr{\'e}, N.~Paragios, and M.~Vakalopoulou, ``Self-supervised nuclei segmentation in histopathological images using attention,'' in \emph{Medical Image Computing and Computer Assisted Intervention--MICCAI 2020: 23rd International Conference, Lima, Peru, October 4--8, 2020, Proceedings, Part V 23}.\hskip 1em plus 0.5em minus 0.4em\relax Springer, 2020, pp. 393--402.

\bibitem{le2022unsupervised}
L.~Le~Bescond, M.~Lerousseau, I.~Garberis, F.~Andr{\'e}, S.~Christodoulidis, M.~Vakalopoulou, and H.~Talbot, ``Unsupervised nuclei segmentation using spatial organization priors,'' in \emph{Medical Image Computing and Computer Assisted Intervention--MICCAI 2022: 25th International Conference, Singapore, September 18--22, 2022, Proceedings, Part II}.\hskip 1em plus 0.5em minus 0.4em\relax Springer, 2022, pp. 325--335.

\bibitem{jiao2006improved}
S.~Jiao, X.~Li, and X.~Lu, ``An improved ostu method for image segmentation,'' in \emph{2006 8th international Conference on Signal Processing}, vol.~2.\hskip 1em plus 0.5em minus 0.4em\relax IEEE, 2006.

\bibitem{liu2020pdam}
D.~Liu, D.~Zhang, Y.~Song, F.~Zhang, L.~O’Donnell, H.~Huang, M.~Chen, and W.~Cai, ``Pdam: A panoptic-level feature alignment framework for unsupervised domain adaptive instance segmentation in microscopy images,'' \emph{IEEE Transactions on Medical Imaging}, vol.~40, no.~1, pp. 154--165, 2020.

\bibitem{hsu2021darcnn}
J.~Hsu, W.~Chiu, and S.~Yeung, ``Darcnn: Domain adaptive region-based convolutional neural network for unsupervised instance segmentation in biomedical images,'' in \emph{Proceedings of the IEEE/CVF conference on computer vision and pattern recognition}, 2021, pp. 1003--1012.

\bibitem{radford2021learning}
A.~Radford, J.~W. Kim, C.~Hallacy, A.~Ramesh, G.~Goh, S.~Agarwal, G.~Sastry, A.~Askell, P.~Mishkin, J.~Clark \emph{et~al.}, ``Learning transferable visual models from natural language supervision,'' in \emph{International conference on machine learning}.\hskip 1em plus 0.5em minus 0.4em\relax PMLR, 2021, pp. 8748--8763.

\bibitem{patashnik2021styleclip}
O.~Patashnik, Z.~Wu, E.~Shechtman, D.~Cohen-Or, and D.~Lischinski, ``Styleclip: Text-driven manipulation of stylegan imagery,'' in \emph{Proceedings of the IEEE/CVF International Conference on Computer Vision}, 2021, pp. 2085--2094.

\bibitem{li2022blip}
J.~Li, D.~Li, C.~Xiong, and S.~Hoi, ``Blip: Bootstrapping language-image pre-training for unified vision-language understanding and generation,'' in \emph{International Conference on Machine Learning}.\hskip 1em plus 0.5em minus 0.4em\relax PMLR, 2022, pp. 12\,888--12\,900.

\bibitem{jain2021putting}
A.~Jain, M.~Tancik, and P.~Abbeel, ``Putting nerf on a diet: Semantically consistent few-shot view synthesis,'' in \emph{Proceedings of the IEEE/CVF International Conference on Computer Vision}, 2021, pp. 5885--5894.

\bibitem{li2022grounded}
L.~H. Li, P.~Zhang, H.~Zhang, J.~Yang, C.~Li, Y.~Zhong, L.~Wang, L.~Yuan, L.~Zhang, J.-N. Hwang \emph{et~al.}, ``Grounded language-image pre-training,'' in \emph{Proceedings of the IEEE/CVF Conference on Computer Vision and Pattern Recognition}, 2022, pp. 10\,965--10\,975.

\bibitem{liu2021medical}
G.~Liu, Y.~Liao, F.~Wang, B.~Zhang, L.~Zhang, X.~Liang, X.~Wan, S.~Li, Z.~Li, S.~Zhang \emph{et~al.}, ``Medical-vlbert: Medical visual language bert for covid-19 ct report generation with alternate learning,'' \emph{IEEE Transactions on Neural Networks and Learning Systems}, vol.~32, no.~9, pp. 3786--3797, 2021.

\bibitem{lu2023visual}
M.~Y. Lu, B.~Chen, A.~Zhang, D.~F. Williamson, R.~J. Chen, T.~Ding, L.~P. Le, Y.-S. Chuang, and F.~Mahmood, ``Visual language pretrained multiple instance zero-shot transfer for histopathology images,'' in \emph{Proceedings of the IEEE/CVF Conference on Computer Vision and Pattern Recognition}, 2023, pp. 19\,764--19\,775.

\bibitem{wu2023zeroshot}
Y.~Wu, Y.~Zhou, J.~Saiyin, B.~Wei, M.~Lai, J.~Shou, Y.~Fan, and Y.~Xu, ``Zero-shot nuclei detection via visual-language pre-trained models,'' in \emph{Medical Image Computing and Computer Assisted Intervention -- MICCAI 2023}, H.~Greenspan, A.~Madabhushi, P.~Mousavi, S.~Salcudean, J.~Duncan, T.~Syeda-Mahmood, and R.~Taylor, Eds.\hskip 1em plus 0.5em minus 0.4em\relax Cham: Springer Nature Switzerland, 2023, pp. 693--703.

\bibitem{qin2022medical}
Z.~Qin, H.~Yi, Q.~Lao, and K.~Li, ``Medical image understanding with pretrained vision language models: A comprehensive study,'' \emph{arXiv preprint arXiv:2209.15517}, 2022.

\bibitem{yamada2022lemons}
Y.~Yamada, Y.~Tang, and I.~Yildirim, ``When are lemons purple? the concept association bias of clip,'' \emph{arXiv preprint arXiv:2212.12043}, 2022.

\bibitem{radford2019language}
A.~Radford, J.~Wu, R.~Child, D.~Luan, D.~Amodei, I.~Sutskever \emph{et~al.}, ``Language models are unsupervised multitask learners,'' \emph{OpenAI blog}, vol.~1, no.~8, p.~9, 2019.

\bibitem{wang2021knowledge}
L.~Wang and K.-J. Yoon, ``Knowledge distillation and student-teacher learning for visual intelligence: A review and new outlooks,'' \emph{IEEE transactions on pattern analysis and machine intelligence}, vol.~44, no.~6, pp. 3048--3068, 2021.

\bibitem{dopido2013semisupervised}
I.~D{\'o}pido, J.~Li, P.~R. Marpu, A.~Plaza, J.~M.~B. Dias, and J.~A. Benediktsson, ``Semisupervised self-learning for hyperspectral image classification,'' \emph{IEEE transactions on geoscience and remote sensing}, vol.~51, no.~7, pp. 4032--4044, 2013.

\bibitem{chen2019catastrophic}
X.~Chen, S.~Wang, B.~Fu, M.~Long, and J.~Wang, ``Catastrophic forgetting meets negative transfer: Batch spectral shrinkage for safe transfer learning,'' \emph{Advances in Neural Information Processing Systems}, vol.~32, 2019.

\bibitem{irshad2013methods}
H.~Irshad, A.~Veillard, L.~Roux, and D.~Racoceanu, ``Methods for nuclei detection, segmentation, and classification in digital histopathology: a review—current status and future potential,'' \emph{IEEE reviews in biomedical engineering}, vol.~7, pp. 97--114, 2013.

\bibitem{li2019signet}
J.~Li, S.~Yang, X.~Huang, Q.~Da, X.~Yang, Z.~Hu, Q.~Duan, C.~Wang, and H.~Li, ``Signet ring cell detection with a semi-supervised learning framework,'' in \emph{Information Processing in Medical Imaging: 26th International Conference, IPMI 2019, Hong Kong, China, June 2--7, 2019, Proceedings 26}.\hskip 1em plus 0.5em minus 0.4em\relax Springer, 2019, pp. 842--854.

\bibitem{qu2020weakly}
H.~Qu, P.~Wu, Q.~Huang, J.~Yi, Z.~Yan, K.~Li, G.~M. Riedlinger, S.~De, S.~Zhang, and D.~N. Metaxas, ``Weakly supervised deep nuclei segmentation using partial points annotation in histopathology images,'' \emph{IEEE transactions on medical imaging}, vol.~39, no.~11, pp. 3655--3666, 2020.

\bibitem{lin2014microsoft}
T.-Y. Lin, M.~Maire, S.~Belongie, J.~Hays, P.~Perona, D.~Ramanan, P.~Doll{\'a}r, and C.~L. Zitnick, ``Microsoft coco: Common objects in context,'' in \emph{Computer Vision--ECCV 2014: 13th European Conference, Zurich, Switzerland, September 6-12, 2014, Proceedings, Part V 13}.\hskip 1em plus 0.5em minus 0.4em\relax Springer, 2014, pp. 740--755.

\bibitem{chen2023vlp}
F.-L. Chen, D.-Z. Zhang, M.-L. Han, X.-Y. Chen, J.~Shi, S.~Xu, and B.~Xu, ``Vlp: A survey on vision-language pre-training,'' \emph{Machine Intelligence Research}, vol.~20, no.~1, pp. 38--56, 2023.

\bibitem{jaiswal2020survey}
A.~Jaiswal, A.~R. Babu, M.~Z. Zadeh, D.~Banerjee, and F.~Makedon, ``A survey on contrastive self-supervised learning,'' \emph{Technologies}, vol.~9, no.~1, p.~2, 2020.

\bibitem{zhou2022extract}
C.~Zhou, C.~C. Loy, and B.~Dai, ``Extract free dense labels from clip,'' in \emph{European Conference on Computer Vision}.\hskip 1em plus 0.5em minus 0.4em\relax Springer, 2022, pp. 696--712.

\bibitem{zhao2022exploiting}
S.~Zhao, Z.~Zhang, S.~Schulter, L.~Zhao, B.~Vijay~Kumar, A.~Stathopoulos, M.~Chandraker, and D.~N. Metaxas, ``Exploiting unlabeled data with vision and language models for object detection,'' in \emph{European Conference on Computer Vision}.\hskip 1em plus 0.5em minus 0.4em\relax Springer, 2022, pp. 159--175.

\bibitem{lin2022learning}
C.~Lin, P.~Sun, Y.~Jiang, P.~Luo, L.~Qu, G.~Haffari, Z.~Yuan, and J.~Cai, ``Learning object-language alignments for open-vocabulary object detection,'' \emph{arXiv preprint arXiv:2211.14843}, 2022.

\bibitem{eslami2021does}
S.~Eslami, G.~de~Melo, and C.~Meinel, ``Does clip benefit visual question answering in the medical domain as much as it does in the general domain?'' \emph{arXiv preprint arXiv:2112.13906}, 2021.

\bibitem{liu2023clip}
J.~Liu, Y.~Zhang, J.-N. Chen, J.~Xiao, Y.~Lu, B.~A. Landman, Y.~Yuan, A.~Yuille, Y.~Tang, and Z.~Zhou, ``Clip-driven universal model for organ segmentation and tumor detection,'' \emph{arXiv preprint arXiv:2301.00785}, 2023.

\bibitem{liu2023pre}
P.~Liu, W.~Yuan, J.~Fu, Z.~Jiang, H.~Hayashi, and G.~Neubig, ``Pre-train, prompt, and predict: A systematic survey of prompting methods in natural language processing,'' \emph{ACM Computing Surveys}, vol.~55, no.~9, pp. 1--35, 2023.

\bibitem{wei2022chain}
J.~Wei, X.~Wang, D.~Schuurmans, M.~Bosma, F.~Xia, E.~Chi, Q.~V. Le, D.~Zhou \emph{et~al.}, ``Chain-of-thought prompting elicits reasoning in large language models,'' \emph{Advances in Neural Information Processing Systems}, vol.~35, pp. 24\,824--24\,837, 2022.

\bibitem{zhang2022automatic}
Z.~Zhang, A.~Zhang, M.~Li, and A.~Smola, ``Automatic chain of thought prompting in large language models,'' \emph{arXiv preprint arXiv:2210.03493}, 2022.

\bibitem{lester2021power}
B.~Lester, R.~Al-Rfou, and N.~Constant, ``The power of scale for parameter-efficient prompt tuning,'' \emph{arXiv preprint arXiv:2104.08691}, 2021.

\bibitem{brown2020language}
T.~Brown, B.~Mann, N.~Ryder, M.~Subbiah, J.~D. Kaplan, P.~Dhariwal, A.~Neelakantan, P.~Shyam, G.~Sastry, A.~Askell \emph{et~al.}, ``Language models are few-shot learners,'' \emph{Advances in neural information processing systems}, vol.~33, pp. 1877--1901, 2020.

\bibitem{gu2023systematic}
J.~Gu, Z.~Han, S.~Chen, A.~Beirami, B.~He, G.~Zhang, R.~Liao, Y.~Qin, V.~Tresp, and P.~Torr, ``A systematic survey of prompt engineering on vision-language foundation models,'' \emph{arXiv preprint arXiv:2307.12980}, 2023.

\bibitem{zhou2022learning}
K.~Zhou, J.~Yang, C.~C. Loy, and Z.~Liu, ``Learning to prompt for vision-language models,'' \emph{International Journal of Computer Vision}, vol. 130, no.~9, pp. 2337--2348, 2022.

\bibitem{jia2022visual}
M.~Jia, L.~Tang, B.-C. Chen, C.~Cardie, S.~Belongie, B.~Hariharan, and S.-N. Lim, ``Visual prompt tuning,'' in \emph{European Conference on Computer Vision}.\hskip 1em plus 0.5em minus 0.4em\relax Springer, 2022, pp. 709--727.

\bibitem{hinton2015distilling}
G.~Hinton, O.~Vinyals, and J.~Dean, ``Distilling the knowledge in a neural network,'' \emph{arXiv preprint arXiv:1503.02531}, 2015.

\bibitem{buciluǎ2006model}
C.~Buciluǎ, R.~Caruana, and A.~Niculescu-Mizil, ``Model compression,'' in \emph{Proceedings of the 12th ACM SIGKDD international conference on Knowledge discovery and data mining}, 2006, pp. 535--541.

\bibitem{ahn2019variational}
S.~Ahn, S.~X. Hu, A.~Damianou, N.~D. Lawrence, and Z.~Dai, ``Variational information distillation for knowledge transfer,'' in \emph{Proceedings of the IEEE/CVF conference on computer vision and pattern recognition}, 2019, pp. 9163--9171.

\bibitem{zhang2023vision}
J.~Zhang, J.~Huang, S.~Jin, and S.~Lu, ``Vision-language models for vision tasks: A survey,'' \emph{arXiv preprint arXiv:2304.00685}, 2023.

\bibitem{ding2022decoupling}
J.~Ding, N.~Xue, G.-S. Xia, and D.~Dai, ``Decoupling zero-shot semantic segmentation,'' in \emph{Proceedings of the IEEE/CVF Conference on Computer Vision and Pattern Recognition}, 2022, pp. 11\,583--11\,592.

\bibitem{luddecke2022image}
T.~L{\"u}ddecke and A.~Ecker, ``Image segmentation using text and image prompts,'' in \emph{Proceedings of the IEEE/CVF Conference on Computer Vision and Pattern Recognition}, 2022, pp. 7086--7096.

\bibitem{wang2023object}
L.~Wang, Y.~Liu, P.~Du, Z.~Ding, Y.~Liao, Q.~Qi, B.~Chen, and S.~Liu, ``Object-aware distillation pyramid for open-vocabulary object detection,'' in \emph{Proceedings of the IEEE/CVF Conference on Computer Vision and Pattern Recognition}, 2023, pp. 11\,186--11\,196.

\bibitem{chen2021crossvit}
C.-F.~R. Chen, Q.~Fan, and R.~Panda, ``Crossvit: Cross-attention multi-scale vision transformer for image classification,'' in \emph{Proceedings of the IEEE/CVF international conference on computer vision}, 2021, pp. 357--366.

\bibitem{kumar2017dataset}
N.~Kumar, R.~Verma, S.~Sharma, S.~Bhargava, A.~Vahadane, and A.~Sethi, ``A dataset and a technique for generalized nuclear segmentation for computational pathology,'' \emph{IEEE transactions on medical imaging}, vol.~36, no.~7, pp. 1550--1560, 2017.

\bibitem{antol2015vqa}
S.~Antol, A.~Agrawal, J.~Lu, M.~Mitchell, D.~Batra, C.~L. Zitnick, and D.~Parikh, ``Vqa: Visual question answering,'' in \emph{Proceedings of the IEEE international conference on computer vision}, 2015, pp. 2425--2433.

\bibitem{he2016automatic}
Y.~He, K.~Chakrabarti, T.~Cheng, and T.~Tylenda, ``Automatic discovery of attribute synonyms using query logs and table corpora,'' in \emph{Proceedings of the 25th International Conference on World Wide Web}, 2016, pp. 1429--1439.

\bibitem{ge2021yolox}
Z.~Ge, S.~Liu, F.~Wang, Z.~Li, and J.~Sun, ``Yolox: Exceeding yolo series in 2021,'' \emph{arXiv preprint arXiv:2107.08430}, 2021.

\bibitem{he2017mask}
K.~He, G.~Gkioxari, P.~Doll{\'a}r, and R.~Girshick, ``Mask r-cnn,'' in \emph{Proceedings of the IEEE international conference on computer vision}, 2017, pp. 2961--2969.

\bibitem{cheplygina2019not}
V.~Cheplygina, M.~de~Bruijne, and J.~P. Pluim, ``Not-so-supervised: a survey of semi-supervised, multi-instance, and transfer learning in medical image analysis,'' \emph{Medical image analysis}, vol.~54, pp. 280--296, 2019.

\bibitem{qu2019weakly}
H.~Qu, P.~Wu, Q.~Huang, J.~Yi, G.~M. Riedlinger, S.~De, and D.~N. Metaxas, ``Weakly supervised deep nuclei segmentation using points annotation in histopathology images,'' in \emph{International Conference on Medical Imaging with Deep Learning}.\hskip 1em plus 0.5em minus 0.4em\relax PMLR, 2019, pp. 390--400.

\bibitem{qu2020nuclei}
H.~Qu, J.~Yi, Q.~Huang, P.~Wu, and D.~Metaxas, ``Nuclei segmentation using mixed points and masks selected from uncertainty,'' in \emph{2020 IEEE 17th International Symposium on Biomedical Imaging (ISBI)}.\hskip 1em plus 0.5em minus 0.4em\relax IEEE, 2020, pp. 973--976.

\bibitem{liu2022weakly}
W.~Liu, Q.~He, and X.~He, ``Weakly supervised nuclei segmentation via instance learning,'' in \emph{2022 IEEE 19th International Symposium on Biomedical Imaging (ISBI)}.\hskip 1em plus 0.5em minus 0.4em\relax IEEE, 2022, pp. 1--5.

\bibitem{wang2022freesolo}
X.~Wang, Z.~Yu, S.~De~Mello, J.~Kautz, A.~Anandkumar, C.~Shen, and J.~M. Alvarez, ``Freesolo: Learning to segment objects without annotations,'' in \emph{Proceedings of the IEEE/CVF Conference on Computer Vision and Pattern Recognition}, 2022, pp. 14\,176--14\,186.

\bibitem{chen2022unsupervised}
P.~Chen, C.~Zhu, Z.~Shui, J.~Cai, S.~Zheng, S.~Zhang, and L.~Yang, ``Unsupervised dense nuclei detection and segmentation with prior self-activation map for histology images,'' \emph{arXiv preprint arXiv:2210.07862}, 2022.

\bibitem{wang2023cut}
X.~Wang, R.~Girdhar, S.~X. Yu, and I.~Misra, ``Cut and learn for unsupervised object detection and instance segmentation,'' in \emph{Proceedings of the IEEE/CVF Conference on Computer Vision and Pattern Recognition}, 2023, pp. 3124--3134.

\bibitem{achanta2012slic}
R.~Achanta, A.~Shaji, K.~Smith, A.~Lucchi, P.~Fua, and S.~S{\"u}sstrunk, ``Slic superpixels compared to state-of-the-art superpixel methods,'' \emph{IEEE transactions on pattern analysis and machine intelligence}, vol.~34, no.~11, pp. 2274--2282, 2012.

\bibitem{shi2022nuclei}
P.~Shi, J.~Zhong, L.~Lin, L.~Lin, H.~Li, and C.~Wu, ``Nuclei segmentation of he stained histopathological images based on feature global delivery connection network,'' \emph{Plos one}, vol.~17, no.~9, p. e0273682, 2022.

\end{thebibliography}
\end{document}